# Leveraging Teleconnections with Physics-Informed Graph Attention Networks for Long-Range Extreme Rainfall Forecasting in Thailand

Kiattikun Chobtham[a], Kanoksri Sarinnapakorn[a], Kritanai Torsri[a],
Prattana Deeprasertkul[a], Jirawan Kamma[a]

[a]Hydro-Informatics Institute, Ministry of Higher Education, Science, Research and Innovation, Bangkok 10900, Thailand

## Abstract

Accurate rainfall forecasting, particularly for extreme events, remains a significant challenge in climatology and the Earth system. This paper presents a novel physics-informed Graph Neural Network (GNN) combined with extreme-value analysis techniques to improve gauge-station rainfall predictions across Thailand. The model leverages a graph-structured representation of gauge stations to capture complex spatiotemporal patterns, and it offers explainability through teleconnections. We preprocess relevant climate indices that potentially influence regional rainfall. The proposed Graph Attention Network (GAT) with a detrending technique (GAT-detrend) applies the attention mechanism using initial edge features derived from simple orographic-precipitation physics. The detrended non-stationary time series and embeddings are processed by GAT-detrend. To address extremes, we perform Peak-Over-Threshold (POT) mapping using the novel Spatial Season-aware Generalized Pareto Distribution (GPD) method, which overcomes limitations of traditional machine-learning models. Experiments demonstrate that our method outperforms well-established baselines across most regions, including areas prone to extremes, and achieves state-of-the-art accuracy for extreme rainfall. Compared with the operational forecasting system SEAS5, our real-world application reduces average monthly RMSE by 21.83 mm/month and raises rainfall accuracy from 72.1% to 79.3% against ground truth, producing high-resolution maps that support decision-making in long-term water management. This illustrates the practical value for other regions with monsoon-driven variability.

## 1 Introduction

Climate change is making extreme weather more frequent and severe, and hence improving climate prediction remains challenging. Several studies in Numerical Weather Prediction (NWP) and Machine Learning (ML) have made significant advancements in climate modeling for short-term (one hour-15 days) or medium-term (2-6 weeks of sub-seasonal forecasting) climate predictions. Classical dynamic weather forecasting models or NWP typically employ computer simulations for short-term forecasting based on initial weather variables. ML models are potentially less computationally demanding at inference, and deep learning can uncover complex nonlinear relationships and accurately produce weather predictions. Recent ML studies include Graph Neural Networks (GNNs) (Scarselli et al., 2018), and Physics-Informed Neural Networks (PINNs) (Raissi et al., 2019), which can excel in prediction tasks. In terms of global short-term weather prediction, FourCastNet (Kurth et al., 2023; Bonev et al., 2025), GraphCast (Lam et al., 2023), Aardvark Weather (Allen et al., 2025), and Pangu-Weather (Bi et al., 2023) were introduced as compelling alternatives to NWP models. Gong et al. (2024) proposed a spatio-temporal learning framework operating at the station level to improve short-term weather forecasting skill. The novel deep-learning model ClimODE (Verma et al., 2024) was proposed to learn a spatiotemporal continuous-time process governed by atmospheric physics. However, the ML models remain limited to short-term predictions due to error propagation, and many models do not predict precipitation.

Long-term climate prediction models produce forecasts months in advance to address the challenges of water management in reservoirs and agricultural planning. This forecasting is feasible because some climate features within the Earth system evolve gradually. Their slower patterns are predictable over extended timescales. For example, teleconnections from Sea Surface Temperatures (SSTs) in the Niño 1+2, Niño 3 and Niño 4 regions (El Niño–Southern Oscillation climate index: ENSO) in the Pacific Ocean or the North Pacific area (Pacific Decadal Oscillation climate index: PDO) in specific years can significantly affect rainfall patterns in various areas during El Niño or La Niña events. The study (Fuentes-Franco et al., 2016) provided insights into the statistical analysis of climate variables or climate indices, by examining historical trends and precipitation patterns at both global and regional scales in many regions.

In the northern peninsular Malaysia area, ENSO exerts a strong influence on the timing and intensity of the monsoon (Moten et al., 2014), affecting rainfall distribution and the apparent length of the rainy season. Specifically in Thailand (Dechpichai et al., 2022), the hydroclimate is governed by a monsoon regime. The principal rainy or wet season spans mid-May to mid-October during the Southwest monsoon, which advects humid air from the Indian Ocean and produces widespread heavy rainfall. In contrast, the Northeast monsoon transports colder and drier continental air from China into Thailand from October to December. The transition period from March to May is typically marked by rising temperatures as the region shifts from the Northeast to the Southwest monsoon, known as the monsoon onset period (Khedari et al., 2002; Chokngamwong & Chiu, 2006). However, the rainfall patterns in southern Thailand differ from those in other regions (Amnuaylojaroen, 2021). This is because the region is near the Intertropical Convergence Zone (ITCZ) and is impacted by the eastward-propagating wind patterns of the Madden-Julian Oscillation (MJO) and Boreal Summer Intraseasonal Oscillation (BSISO) (Chang et al. 2021).

SEAS5 (Johnson et al., 2019) is the fifth-generation seasonal forecasting system and a widely used public benchmark developed by the European Centre for Medium-Range Weather Forecasts (ECMWF). It couples the atmosphere of the Integrated Forecasting System (IFS) to ocean and sea-ice components with 91 vertical levels to produce probabilistic, long-range seasonal precipitation forecasts. The operational product uses a perturbation scheme to generate a 51-member ensemble and issues monthly global seasonal outlooks at roughly 36-km horizontal resolution. However, these methods are often time-consuming and require substantial expert knowledge, potentially imposing strong mechanistic assumptions on the Earth system. Consequently, interest has shifted toward data-driven approaches based on ML methods. However, interpretability is an ongoing difficulty, as causal mechanisms are often difficult to extract from ML black-box models. As a result, these models hinder practitioners from making decisions in real-world situations (Waqas et al., 2025). Methods from explainable AI and causal discovery algorithms (Kitson et al., 2023) can help elucidate the relative contributions of various climate drivers and their interactions in modulating climate (Runge et al., 2019; Du et al., 2024), but a causal model must be represented by a directed acyclic graph and time-series prediction remains particularly challenging (Ravuri et al., 2021). Research recommendations (Saha & Ravela, 2024; Camps-Valls et al., 2023) emphasize that data-driven ML and traditional physical modeling can complement each other in predicting extreme events. In addition, a clear understanding of the Earth system is a critical step for developing accurate predictive models.

In this work, we hypothesize that embedding physics as edge features in graph-structured GNNs improves spatiotemporal learning over purely data-driven GNNs. We cast the long-term rainfall prediction problem for a region of Thailand into a graph-structured model that offers explainability through learned teleconnections. We enhance the attention mechanism of standard GATs to learn from non-stationary time-series data using a physics-informed framework, GAT-detrend, which automatically weights the contribution of climate features and their spatial teleconnections. Moreover, we propose a novel Spatial Season-aware GPD method for extreme-value analysis, paired with GAT-detrend, to improve extreme rainfall prediction.

Our work consists of the following steps:

1) Construct and reanalyse graph-structured models based on physics-inspired initial edge features derived from orographic precipitation (details are given in Subsection 2.1), clusters of gauge stations and feature selection for each cluster.
2) Learn nonlinear relationships from spatiotemporal data with the proposed GAT-detrend models for each cluster of gauge stations (details are given in Section 3).
3) Improve extreme rainfall prediction by mapping the predicted values with POT using the novel Spatial Season-aware GPD method (details are given in Subsection 4.1).

The paper is organised as follows: Section 2 describes related work; Sections 3 and 4 present physics-informed GAT-detrend and Spatial Season-aware GPD, respectively; Section 5 describes the dataset and experimental setup; Section 6 presents experimental results; and Section 7 provides concluding remarks.

## 2 Related Work

### 2.1 Orographic Precipitation

As mentioned in the introduction, various studies suggest integrating physics with data-driven ML. The simple thermodynamic equation of Smith linear model (Smith, 2003), governed by the physics of orographic precipitation, provides the spectral formulation that captures the basic structure of rainfall fields. The formula is applicable to high-mountain meteorology as follows:

$$\hat{P}(k,\mathrm{l}) = \frac{C_w i\sigma\hat{h}(k,\mathrm{l})}{(1+i\sigma\tau_c)(1+i\sigma\,\tau_h)} \quad (1)$$

where $\hat{P}(k,\mathrm{l})$ is the spectral transform of the precipitation field, $C_w$ is the condensation-efficiency constant calculated using air density and saturation specific humidity, $\hat{h}(k,\mathrm{l})$ is the Fourier transform of the terrain elevation, $\tau_c$ is the conversion time scale from cloud water to rain (hydrometeors), $\tau_h$ is the time constant for hydrometeor fallout, k and l are wavenumbers, $\sigma =$

Uk + Vl, U and V are zonal and meridional wind components, and $i$ denotes the imaginary number. The orographic-precipitation formulation is the short-term precipitation governing equation for advection, conversion, and fallout. Precipitation is distributed downwind from each location over a distance equal to the product of the wind speed and the cloud time constant. Figure 1 shows a simple simulation of equation (1) where the wind speed is 40 m/s, $\tau_c$ is 1000, $\tau_h$ is 500, and the mountain height is 2000 m. For long-term climate forecasting, machine-learning models can potentially use the probability of orographic-precipitation based on the monthly average simulations as prior knowledge. Therefore, terrain elevation, the monthly average U and V variables should be included as model features.

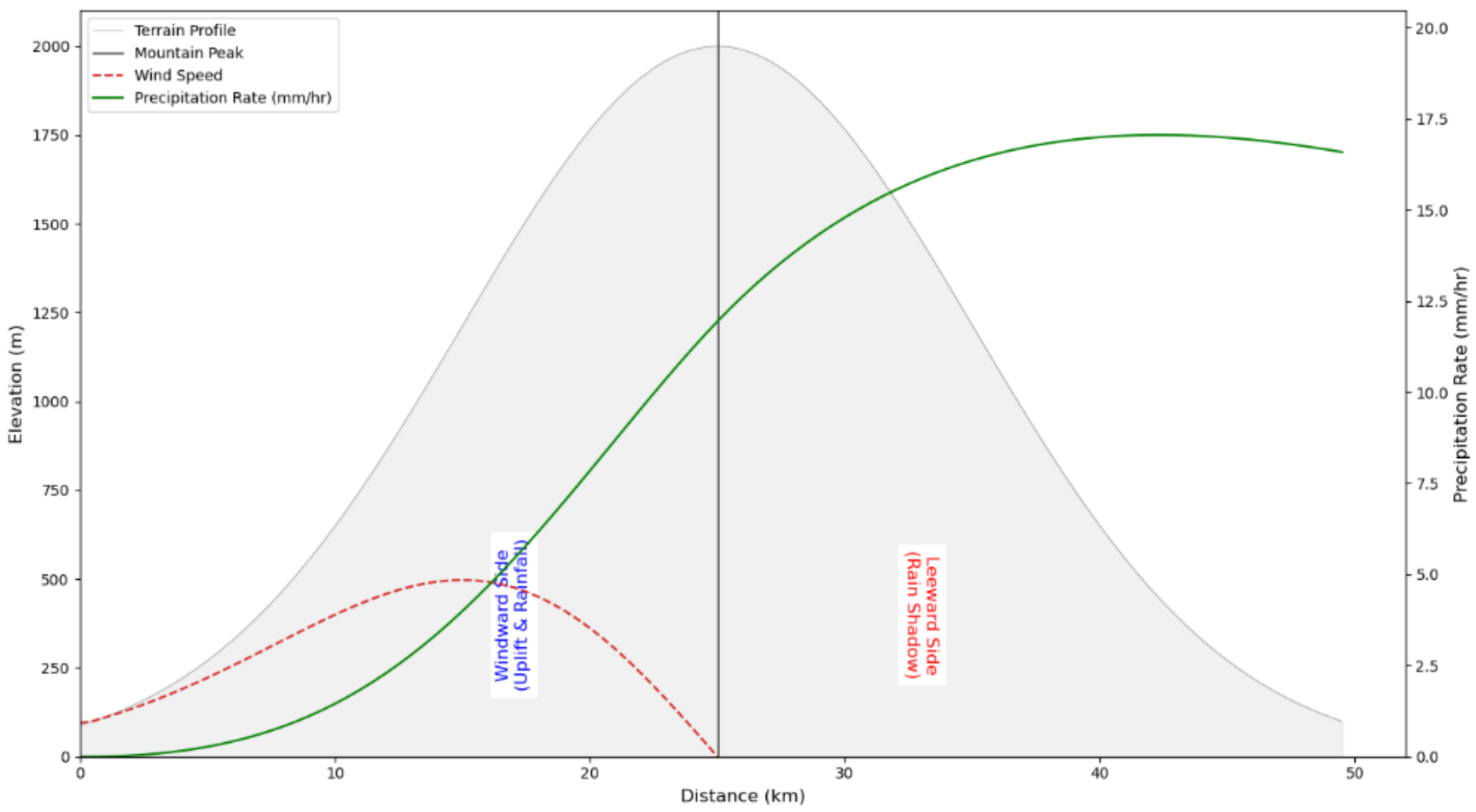

**Figure 1.** Simulation of the simple thermodynamic equation of Smith linear model

## 2.2 Recurrent Neural Networks (RNNs) and Graph Neural Networks (GNNs)

Recent advancements in forecasting long-term climate variability have increasingly relied on ML and deep learning models. Recurrent Neural Networks (RNNs), LSTM, and Gated Recurrent Unit (GRU) were employed for sequence-to-sequence prediction (Hochreiter and Schmidhuber, 1997; Cho et al., 2014). Consequently, hybrid models (Waqas et al., 2024) using biorthogonal discrete wavelet transformation (Hammond et al, 2011) with LSTM (Wavelet-CNN-LSTM and Wavelet-LSTM) showed superior performance in monthly gauge-rainfall predictions compared to traditional models.

Recently, GNNs have become a well-established neural network architecture designed for processing graph-structured data. GNNs can take input in the form of a directed or undirected graph describing the connectivity structure, making them more appropriate for spatial data that lack a regular grid in a non-Euclidean domain. This architecture learns node representations through a message-passing mechanism, followed by aggregation to produce an embedding for the entire graph. Shan et al. (2025) introduced a GNN incorporating physical edge features (edge distance). This bridges physics-based understanding with interpretable AI.

Several approaches using GNNs with spatiotemporal data were proposed to learn graph representations for supervised tasks, although the majority of applications addressed traffic forecasting rather than weather forecasting. For example, Chen et al. (2022) modified Graph Convolution Networks (GCNs) (Kipf & Welling, 2017) and LSTMs and proposed GC-LSTM to predict future links, achieving outstanding performance and outperforming state-of-the-art methods. GCNs were formulated for undirected graphs, operating on the normalised graph

Laplacian. Their convolution is implemented using a localised first-order Chebyshev polynomial of the Laplacian. Pareja et al. (2020) proposed EvolveGCN, which learns node embeddings over time by combining dynamic graphs with RNNs (two variants, EvolveGCN-H and EvolveGCN-O, use GRU and LSTM, respectively) to evolve the weights in GCNs, which serve as the hidden states of the RNN architecture. A Graph Multi-Attention Network (GMAN) was introduced to predict traffic conditions several steps ahead using a Transformer attention layer (Zheng et al., 2019). The model employed both spatial and temporal attention mechanisms to encode traffic features and generate output sequences. Wu et al. (2018) proposed Graph Attention LSTM (GAT-LSTM) to forecast traffic flow as a time-series problem on road networks. They constructed a directed graph of the road network and stacked multiple GAT-LSTM layers to learn from this structure, addressing the link-level traffic-forecasting problem. The GNN weather forecasting framework GraphCast embeds each node in a non-Euclidean domain (Lam et al., 2023), but the model is used for short-term weather forecasting with a 6-hour time-step prediction, and the number of autoregressive steps can be increased up to 3 days. Consequently, this autoregressive prediction can lead to error propagation in long-term weather forecasting. The success of these models suggests they can be effectively leveraged for long-term rainfall forecasting.

While attention and self-attention strategies have proven remarkably successful in sequential domains like Large Language Models (LLMs), Graph Attention Networks (GATs) extend this capability to graph-structured data (Veličković et al., 2018; Brody et al., 2021). As an enhanced variant of the traditional GCN architecture, GATs employ an attention mechanism that allows nodes to dynamically weight the importance of their features during aggregation without requiring prior knowledge of edge features. Because our approach applies GATs to the long-term rainfall-forecasting problem to mitigate the limitations of static weights and error propagation, we introduce necessary preliminary information. Formally, we define static and dynamic graphs as follows:

A static graph $\mathrm{G} = (\mathrm{V}, \mathrm{E})$ consists of a set of nodes $\mathrm{V}$ and a set of edges $\mathrm{E}$, while dynamic graphs can be represented as snapshot graphs at time $\mathrm{t}$ where $\mathrm{G_t} = (\mathrm{V}, \mathrm{E_t}); \mathrm{t} \in [1, \mathrm{T}]$.

## 3 GAT-detrend

We propose GAT-detrend, a novel model that integrates spatial teleconnections and attention mechanisms with prior knowledge of edge features ($f_{ij}$) encoding the probability of orographic precipitation. A key limitation of GNNs, however, is the absence of a ground-truth graph structure or a dynamic data-generating process representing the Earth system. We therefore assume a static graph structure over time, while the learned attention coefficients ($\alpha_{\mathrm{ij}}$) dynamically adjust the contribution of each node feature and physics-informed edge feature. This allows more important nodes to receive higher weights during the neighborhood aggregation step.

To enable message passing with temporal awareness, we extend GAT-detrend with two components: a detrending technique to address non-stationary time-series data, and node embeddings that uniquely encode station and climate index identity. Layer Normalization (LayerNorm) plays a critical role in the expressivity of the attention mechanism (Brody et al., 2023); specifically, the derivatives of the mean and variance in LayerNorm serve to re-center and re-scale backward gradients, respectively (Xu et al., 2019).

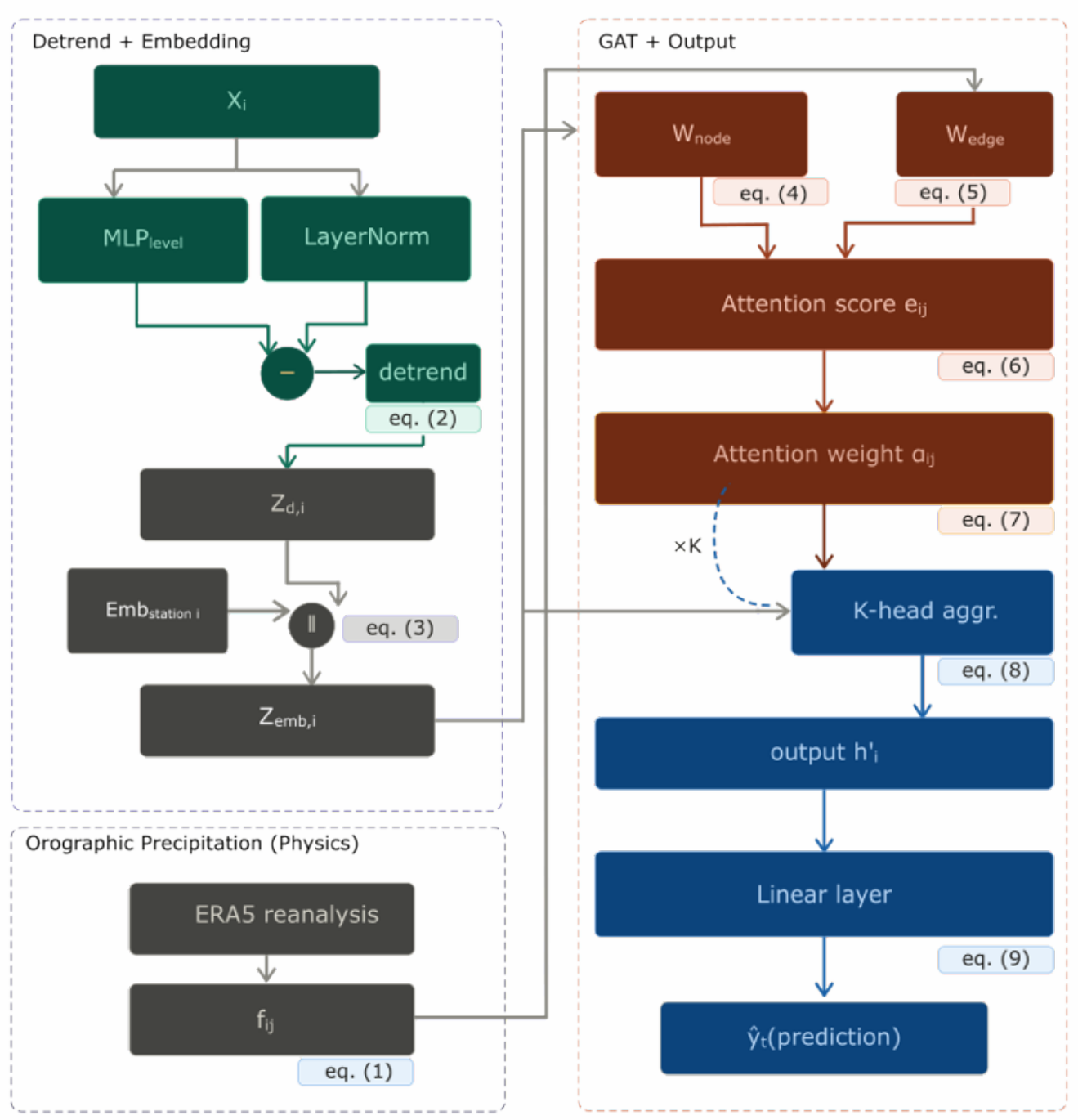

**Figure 2.** The GAT-detrend architecture

Specifically, static edges (E) of graph structures will be constructed using the Pearson correlation and Granger Causality for the time lag (will be described in Subsection 3.2) as the statistical method referred to as teleconnection. This serves as feature selection for data-preprocessing. The GAT-detrend model removes a trend, appends station identity, performs graph attention over the station graph, and projects the graph output to the final hidden representation. The framework is shown in Figure 2, and the formulas of GAT-detrend are:

$$\mathbf{Z_{d,i}} = \mathbf{LayerNorm}(\mathbf{X_{\,i}} - \mathbf{level_i})\,;\mathbf{level_i} = \mathbf{MLP_{level}}(\mathbf{X_{\,i}}) \qquad (2)$$

$$\boldsymbol{Z_{emb,i}} = [\mathbf{Z_{d,i}} || \boldsymbol{Emb_{station,i}}] \quad (3)$$

$$\boldsymbol{z_i} = \boldsymbol{W_{node}} * \boldsymbol{Z_{emb\,,i}} \qquad (4)$$

$$\boldsymbol{g_{ij}} = \boldsymbol{W_{edge}} * \boldsymbol{f_{ij}} \qquad (5)$$

$$\boldsymbol{e_{ij}} = \boldsymbol{\vec{a}^T LeakyReLU}\left(\boldsymbol{z_i}||\boldsymbol{z_j}||\boldsymbol{g_{ij}}\right) \quad (6)$$

$$\boldsymbol{\alpha_{ij}} = \frac{\boldsymbol{exp(e_{ij})}}{\sum_{k\in\mathcal{N}_i}\boldsymbol{exp(e_{ik})}} \quad (7)$$

$$\boldsymbol{\acute{h}_i} = \boldsymbol{\sigma}\left(\frac{\mathbf{1}}{\boldsymbol{K}}\sum_{k=1}^{K}\sum_{j\in\mathcal{N}_i}\boldsymbol{\alpha_{ij}^k\,W^k Z_{emb,j}}\right) \quad (8)$$

$$\boldsymbol{\hat{y}_t} = \boldsymbol{W_{fc}} * \boldsymbol{\acute{h}} + \boldsymbol{b_{fc}} \qquad (9)$$

where $X_i$ is the input node feature at a given time step, LayerNorm denotes Layer Normalization (Xu et al., 2019) to tackle non-stationarity, MLP denotes the feedforward Multi-Layer Perceptron that learns a nonlinear level component, $Z_{d,i} \| Emb_{station,i}$ is the detrended node feature concatenated with a station embedding. || denotes the concatenation operation. $W_{node}$, $W_{edge}$, and $W_{fc}$ are learnable weight matrices. $W^k$ denotes the learnable weight matrix for attention head $k$. $f_{ij}$ is the edge feature between node i to j, $\vec{a}$ is a learnable weight vector, the attention weights ($W_{attention} = \{\boldsymbol{\alpha_{ij}}\}$) represent dynamic attention coefficients given the graph structure using the equation (6)-(7) (Brody et al., 2021). $\mathcal{N}_i$ is a set of neighbors of node i, $\sigma$ denotes the softmax function and $LeakyReLU$ denotes the Leaky Rectified Linear Unit activation function, and $b_{fc}$ denotes the learnable bias term.

Like standard GATs, GAT-detrend can be extended to employ multi-head attention using $K$ independent attention mechanisms to enhance performance where the aggregated features from each head are concatenated or averaged. GAT-detrend explicitly incorporates edge features derived from physics-informed formulations. Specifically, we use the orographic precipitation simulation from Subsection 2.1 as the physics-informed edge feature $f_{ij} = normalisation\ \hat{P}(k, l)$ from Equation (1), which represents the climatological probability of orographic precipitation at the destination station j given its gauge-station elevation at that location (Monthly U and V wind components). Note that $f_{ij}$ does not vary with the location of node i because the graph edges represent teleconnections, not elevation paths between station pairs.

### 3.2 GNN Model Reanalysis

**Algorithm 1:** Hierarchical clustering of rainfall stations

**Input**: monthly rainfall for each station, *latitude* and *longitude* for each station, distance $d$, components $n$ in PCA
**Output**: clusters and their stations
1: ***scaled_rain*** ← Calculate the average monthly rainfall for each station as training data and normalise the data
2: ***data_for_clustering*** ← Reduce dimensionality using PCA with $n$ components given *latitude, longitude* and *scaled_rain*
3: ***clusters*** ← Calculate the Euclidean distance and determine clusters from *data_for_clustering* where the Euclidean distance between cluster centroid < distance $d$

We reanalysed rainfall patterns in Thailand and found that rainfall stations should be grouped into distinct clusters. We use the hierarchical clustering algorithm for rainfall stations in Algorithm 1 and construct static graph structures of GAT-detrend for each cluster using a correlation test with a specific threshold. Algorithm 1 computes the average monthly rainfall for each station over the period, and the values are Z-score normalised. Principal Component Analysis (PCA) is then applied to reduce the data dimensionality using the normalised values and the station coordinates. Next, it computes Euclidean distances among the processed data. The final step of the algorithm performs hierarchical clustering by successively combining clusters whose centroid distances are less than the threshold $d$. Finally, we also incorporate lagged time using the statistical test of Granger Causality, and the test formulation is:

$$Y(t) = \sum_{i=1}^{lag} \alpha_i * Y(t-i) + \sum_{i=1}^{lag} \beta_i * X(t-i) + \epsilon_{y,t} \quad (10)$$

where $\alpha_i$ and $\beta_i$ are coefficients, X(t) denotes the climate index, and Y(t) is rainfall at time t. If the tested p-value is below a significance threshold, we reject the null that lagged X has no effect on Y and conclude that past X affects Y; if p is greater than the significance threshold, we fail to reject the null, indicating insufficient evidence that past X affects Y.

Algorithm 2 describes the training process for GAT-detrend. The algorithm takes as input a static graph and training data with node features, edge, and edge features, along with the number of epochs, the window size, the horizon, and a list of hyperparameters. The list of hyperparameters includes multi-head attention K, number of layers, dropout probability, hidden size, and learning rate to be explored. Algorithm 2 begins by calculating the weight matrix from physics-informed edge features using equation (1). The weight matrices are orographic precipitation probabilities derived from the Smith linear orographic precipitation model. The spectral thermodynamic equation simulates how terrain elevation and wind components (U, V) at each station determine precipitation. These edge features encode the physical process of advection, cloud conversion, and hydrometeor fallout between spatially connected stations. Next, GAT-detrend with its configurations learns attention coefficients on the training dataset. The attention coefficients and graph embeddings are used to compute the predicted values. Finally, the algorithm calculates the loss function from the predicted and true values, and iteratively updates the learnable weights. The model is trained by minimizing the Mean Square Error (MSE) loss.

| **Algorithm 2:** Training algorithm of GAT-detrend | |
|---|---|
| **Input:** | static graph with edge E, features X and target Y, station elevations, U and V wind components, window size W, horizon H, number of epochs N, learning rate r, dropout rate d, multi-head attention k, number of layers l, and hidden size z |
| **Output:** | Trained model |
| 1: | List of snapshots t ← Split signals with window size W and target horizon H given X and Y |
| 2: | $E_{feature}$ ← Simulate physics-informed initial edge features using equation (1) for each station j given U and V wind components, and station elevation |
| 3: | num_epoch = 0 |
| 4: | **While** num_epoch < N **do** |
| 5: | **For** each snapshot t **do** |
| 6: | model ← Define the GAT-detrend model using equations (2) - (9) with $X_t$, E, $E_{feature}$ and hyperparameters d, z, l and k |
| 7: | $Y' \leftarrow \text{model}(X_t, E, E_{feature})$ |
| 8: | Calculate MSE loss ($Y_t'$, $Y_t$) |
| 9: | Backpropagate and update learnable parameters with learning rate r |
| 10: | num_epoch ++ |
| 11: | **Return** model |

## 4 Extreme Value Analysis with Spatial Season-aware GPD

Extreme values are observations that fall in the tails of a distribution with very low probability. In climate prediction, heavy rainfall or heat waves in a short period at a specific location can cause potential catastrophic effects. It is generally well understood that a limitation of deep learning models is that their predicted values often ignore extreme values or the tails of the distribution due to small sample sizes (Sun et al., 2021; Ting, 2024; Camps-Valls et al., 2025). Therefore, we need approaches to specifically tackle extreme-value estimation within the tail region of the relevant distribution. To improve extreme-value accuracy, the state-of-the-art statistical technique Quantile Mapping (QM) is used as a simple bias correction method to align the distribution of model outputs with that of observational data (Maraun, 2016). This method specifically adjusts the Probability

Density Function (PDF) of model outputs to achieve better consistency with the PDF of observed data. However, its accuracy is limited when significant discrepancies exist between model-based and observational datasets. QM is particularly effective for bias correction in reanalysis dataset, especially for precipitation, though it remains unclear whether the mapping function is applicable for long-term forecasting.

In Extreme-Value Theory (EVT), one fits a Generalized Pareto (GP) distribution to rainfall excesses above a properly selected threshold $u$. This approach is also referred to as Peaks-Over-Threshold. The theory suggests that if the Cumulative Distribution Function (CDF) of the scaled excesses $[Y - u | \, Y > u]$ of random variable $Y$ above threshold $u$ converges to a nondegenerate distribution $Q_u$, then this CDF should have the GP form (Lucarini et al., 2016):

$$Q_u(y) = P[\, Y - u < y \,|\, Y > u] = \begin{cases} 1 - \left(1 + \xi \dfrac{(y-u)}{a_u}\right)^{-1/\xi} & ; \xi \neq 0 \\ 1 - e^{-\frac{(y-u)}{a_u}} & ; \xi = 0 \end{cases} \quad (11)$$

where $u$ is the threshold used to determine the excesses of random variable $Y$ ($Y$ denotes rainfall in this study), $\xi$ and $a_u$ are the shape and scale parameters of the distribution.

### 4.1 Spatial Season-aware GPD

We propose a novel approach, Spatial Season-aware GPD, for identifying extreme rainfall values geographically across seasons. Spatial Season-aware GPD maps the ML-predicted rainfall values to the predicted extreme rainfall values given a function learned from observational training data. Because threshold selection is crucial and identifying a threshold for an extreme rainfall value for each station is very difficult, using a percentile automatically adapts the definition of extreme to the conditions (in this study, the percentile threshold is 95). The GPDs from equation (11) are fitted for each station and its corresponding season, based on the spatial location of the stations. The season for each station in the cluster ($i \in Cluster$) in Thailand, as illustrated in Table 1 and Figure 4 (b), was determined by domain experts. To estimate the shape and scale parameters, we use Maximum Likelihood Estimation (MLE) as a commonly used method. During the wet season, predicted values that exceed the threshold with the application of the POT method are mapped using the fitted GP distributions. Note that with very small samples, MLE can be numerically unstable or biased. We fit the GPD only when the number of exceedances exceeds 10, and we impose a hard cap at the maximum observed value to prevent unphysical extrapolation.

| **Cluster** | **Season** | **Season - Month** | **Spatial Season-aware GPD for station $i$** |
|---|---|---|---|
| 1-4 | Wet | SW monsoon (MJJAS), NE monsoon (OND) | $Q_{i,SW}(y)$, $Q_{i,NE}(y)$ |
| | Dry | Monsoon retreat (JFMA) | $Q_{i,dry}(y)$ |
| 5-12 | Wet | Monsoon onset (MAM), Monsoon peak (JJASO) | $Q_{i,onset}(y)$, $Q_{i,peak}(y)$ |
| | Dry | Monsoon retreat (NDJF) | $Q_{i,dry}(y)$ |

**Table 1**. Seasonal classifications and application of Spatial Season-aware GPD by station $i$ in Thailand ($i \in Cluster$)

Finally, the workflow of a real-world application of long-range rainfall forecasting based on physics, GAT-detrend and Spatial Season-aware GPD mapping is shown in the end-to-end forecasting system in Figure 3. Data preprocessing retrieves rainfall from stations along with climate features, performs missing-data imputation, calculates climate indices, and simulates the physics-based estimates of orographic rainfall as edge features for GAT-detrend. Hierarchical clustering of rainfall stations and time-lag processing are performed to construct graph-structured models for each cluster. The graph structures and physics-informed edge features and processed tensor data are used to train GAT-detrend on GPUs. Hyperparameters are fine-tuned to meet target accuracy, followed by Spatial Season-aware GPD extreme-value mapping of the predicted values. For inference, the deployed models are run on a High-Performance Computing (HPC) to generate interpolated maps with 12-month rainfall predictions.

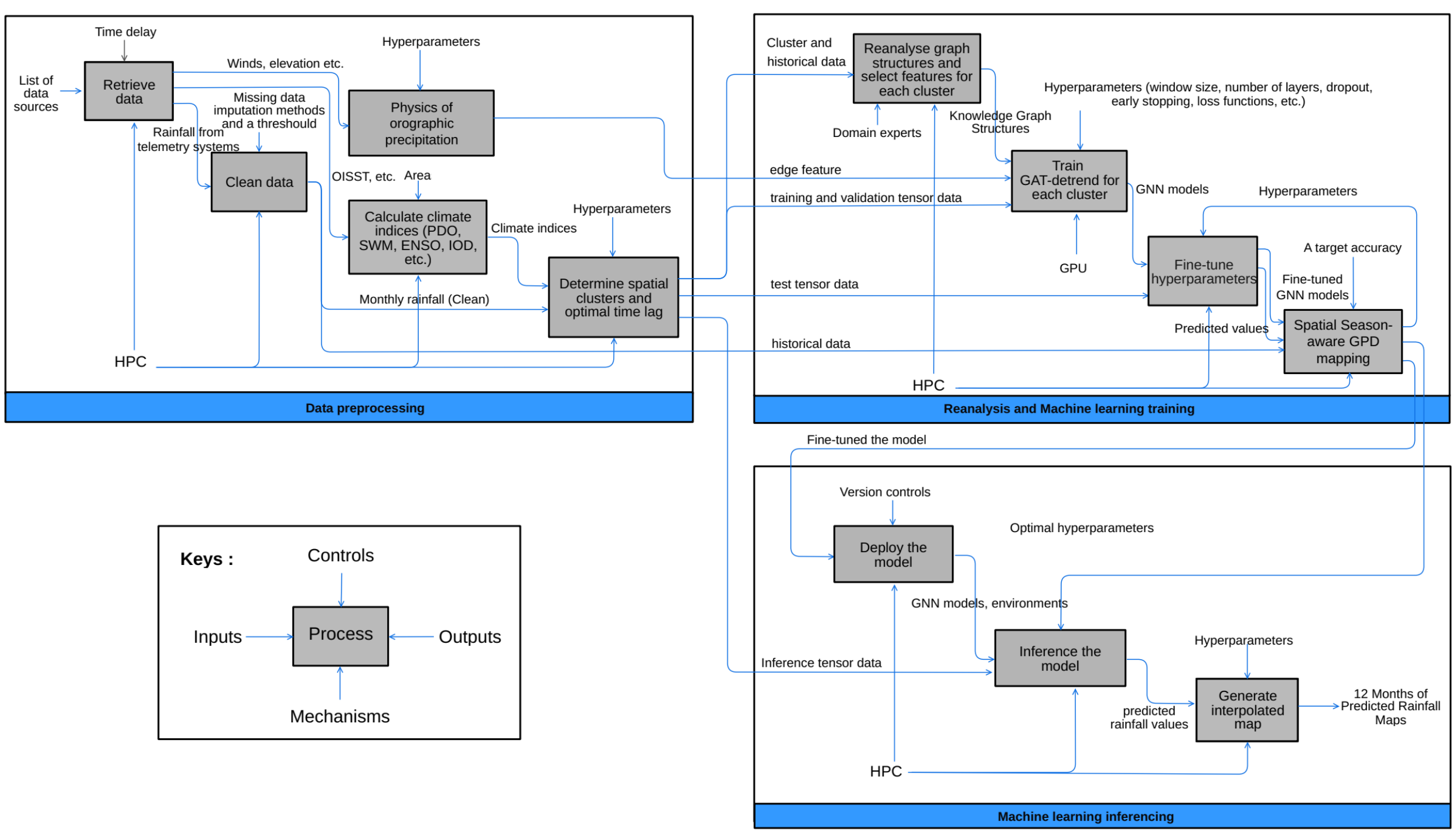

**Figure 3.** Workflow of the operational long-range rainfall forecasts based on physics-informed GAT-detrend and Spatial Season-aware GPD extreme-value mapping.

## 5 Dataset and Experimental Setup

### 5.1 Dataset and Graph Structure Modeling

We use eight climate indices detailed in Table 2, processed from NOAA OISST (Huang et al. 2021), together with historical rainfall data from the Hydro-Informatics Institute (HII) from 2014-2024 as inputs to the hierarchical clustering algorithm (Algorithm 1). Since MJO and BSISO are daily indices, we use the maximum amplitude to obtain the corresponding monthly values. The large number of HII stations covers most areas of Thailand, and hence the dataset is appropriate for clustering. Thai Meteorological Department (TMD) gauge stations from 1982-2024 are used as node features for GAT-detrend. We also encode monthly periodicity of time t with trigonometric time embeddings as follows $\{sin(\frac{2\pi t}{12}), cos(\frac{2\pi t}{12})\}$. Note that forecasts of HII gauge stations are excluded from this experiment because of their limited sample size.

| Climate feature | Description |
|---|---|
| Optimum Interpolation Sea Surface Temperature (OISST) | NOAA long-term climate data record (Huang et al. 2021) blends observations from satellites, ships, buoys, and Argo floats onto a 0.25° × 0.25° global grid. Missing values are interpolated to produce a spatially complete SST map. |
| U and V wind components | U is the eastward (zonal) wind component and V is the northward (meridional) wind component, both in metres per second, where negative values indicate westward and southward flow, respectively. Data are from the ERA5 monthly mean 10m wind components on a regular 0.25° × 0.25° global grid. |
| Pacific Decadal Oscillation (PDO) | The SST Anomaly (SSTA) pattern represents Pacific climate variability. The PDO index is defined as projections of monthly mean SST anomalies onto their first Empirical Orthogonal Function (EOF) vectors in the North Pacific. The North Pacific area is approximately covered by: 20°N-65°N, 120°E-100°W. |
| Indian Ocean Dipole Mode Index (DMI) | The SSTA difference between the western (50°E-70°E, 10°S-10°N) and southeastern (90°E-110°E, 10°S to equator) regions of the tropical Indian Ocean is weighted by latitude. |
| Oceanic Niño Index (ONI) | The SSTA averaged in the east-central equatorial Pacific Ocean (Niño 3.4: 5°S - 5°N, 170°W - 120°W). |
| South West Monsoon Index (SWM) | The calculated SST difference between area A (87°E-91°E, 20°N-21°N) and area B (90°E-95°E, 8°S-10°S) in the tropical Indian Ocean. |
| North East Monsoon Index (NEM) | The optimised SST difference between area A (15.0-21.25°N, 111-119.75°E; 13.75-15°N, 109.75-117.25°E; 12.5-13.75°N, 108.5-116°E) and area B (9.75-12.25°N, 100.25-102.75°E) given initial regions (Moten et al., 2014) in the South China Sea and Gulf of Thailand (Chobtham, K., 2026). |
| Real-time Multivariate MJO series (RMM1) | The major fluctuation in tropical weather occurs on weekly to monthly timescales. The MJO can be characterized as an eastward-moving pulse of cloud and rainfall near the equator that typically recurs every 30 to 60 days. It is calculated using multivariate EOF (MV-EOF) of Outgoing Longwave Radiation (OLR) and zonal wind at 850 hPa (U850), which identifies two modes of interannual variability of the MJO. |
| Boreal Summer Intraseasonal Oscillation (BSISO) | BSISO strongly modulates sub-seasonal to seasonal rainfall variability. It refers to MV-EOF analysis of daily anomalies of OLR and zonal wind at 850 hPa (U850) in the region 10°S-40°N, 40°-160°E, for the extended boreal summer (May-October) season over the 30-year period 1981-2010. |

**Table 2.** Climate features used in the study

Following the spectral method of orographic precipitation (Subsection 2.1), we simulate orographic rainfall using each TMD gauge station elevation and ERA5 monthly average U and V wind components at 10m for 1982–2024. These serve as the GAT-detrend physics-informed initial edge features. For this simulation, we assume that $\tau_c = \tau_h = 1{,}000$s (Jiang and Smith 2003; Smith et al. 2003). We approximate the slope magnitude using each station elevation using Thailand terrain gradients. To construct the GAT-detrend edges, we manually define a static graph structure, and some examples are shown in Figure 5. Each climate feature influences rainfall at individual stations through spatial teleconnections and temporal relationships. The selection of climate indices is based on feature-selection criteria using Pearson correlation. We construct a teleconnection when the average absolute correlation is higher than 0.4. We incorporate domain knowledge into the GAT-detrend edges because some climate indices evolve gradually. For example, the phase of the PDO typically evolves slowly. Therefore, feature selection based solely on Pearson correlation is not reliable for PDO; instead, hard constraints from domain experts are applied to ensure that PDO influence on ENSO indices (e.g., ONI) is properly reflected (Fuentes-Franco et al., 2016). Moreover, we incorporate lagged effects for each climate feature using the Granger Causality test (Granger, 1969) of equation (10) evaluated with the $Chi^2$ test. We consider a teleconnection present when the test p-value is below a significance threshold of 0.1, indicating that the lagged influence of the teleconnection improves each station-level rainfall predictability for each cluster in Thailand.

To perform long-term rainfall forecasting (1–12 months ahead), missing rainfall values for each station are imputed using the monthly median. We exclude TMD stations that report less than 80% of their data for each month across the study period; as a result, 74 TMD stations are retained for the experiments. The full dataset (1982–2024) in Table 3 is split into two folds because the monthly rainfall sample size is very limited and the training cost for all hyperparameter combinations (grid search) is computationally expensive. The 2025 data is allocated for out-of-sample for the operation and the visualization in Subsection 6.4. Final performance is reported by averaging evaluation metrics across both test sets, stations, and horizons, reducing the risk of overfitting to specific La Niña (2021) or El Niño (2024) events within particular time windows. All HII stations in Thailand are clustered and mapped to the locations of the TMD stations; each TMD cluster is then used to train a single model per cluster, rather than training separate models for each station. Therefore, parallel training of all models offers a substantial reduction in computational time.

| Stations | The clusters and High-Quality (HQ) stations | Training, validation data and test data |
|---|---|---|
| TMD Station (132) | All HQ station: (74) | Fold 1:<br>Training data (1982-2019): 456 months<br>Validation data (2020): 12 months<br>Test data: (2021): 12 months<br><br>Fold 2:<br>Training data (1982-2022): 492 months<br>Validation data (2023): 12 months<br>Test data: (2024): 12 months<br><br>Out-of-sample data (2025): 6 months |

**Table 3.** Training, validation data, and test data used in the study

## 5.2 Experimental Setup

We evaluate six models: GAT-detrend without physics or Spatial Season-aware GPD mapping denoted as GAT-detrend (without physics and GPD), GAT-detrend, LSTM, Wavelet-LSTM (Waqas et al., 2024), GC-LSTM (Chen et al., 2022), and EvolveGCN (EvolveGCN-O) (Pareja et al., 2020) implemented in PyTorch and PyTorch Geometric Temporal, with Python 3.8. All experiments are conducted on a HPC system equipped with a 32-core Intel Xeon CPU at 2.3 GHz, 64 GB of RAM, and six NVIDIA V100 GPUs.

In this experiment, we use a 24-month input window and a 12-month prediction horizon throughout the paper. Training hyperparameters are as follows: hidden layer sizes = [16, 32, 128]; the number of epochs = 300 with early stopping (patience=50) on the validation set; the number of GAT-detrend multi-heads = [1, 4, 8]; the number of layers = [1, 2]; the L2 regularization of parameters is set to 0.0001 and the dropout rate = [0, 0.2, 0.5]. To tackle non-stationary time series, we use a station embedding technique that encodes each node identity, where station embedding dimension = 8 and SiLU activations of MLP estimates a level signal for a local trend. Chebyshev filter sizes for GC-LSTM and EvolveGCN = [2, 3, 4, 5]. We use the bior1.3 wavelet at level 2 as the decomposition function for Wavelet-LSTM which outperforms other setting (Waqas et al., 2024). All models are trained with the MSE loss with a learning rate of 0.01.

### 5.3 Evaluation Metrics

We use three evaluation metrics: Root Mean Squared Error (RMSE), Nash–Sutcliffe Efficiency (NSE), and accuracy to assess the performance of each approach. RMSE is calculated as:

$$RMSE = \sqrt[2]{MSE}\ , MSE = \frac{\sum (y_{i,t} - \hat{y}_{i,t})^2}{n}$$

Accuracy and Mean Absolute Percentage Error (MAPE) are calculated as follows:

$$Accuracy = 100 - MAPE = 100 - \frac{\sum\left(\left(\left|y_{i,t} - \hat{y}_{i,t}\right| / \ y_{i,t}\right)\right)}{n} \times 100$$

Because MAPE is undefined when $y_{i,t}$=0, we use the Symmetric Mean Absolute Percentage Error (SMAPE), defined as:

$$SMAPE = \frac{\sum\left(2\frac{\left|y_{i,t} - \hat{y}_{i,t}\right|}{\left|y_{i,t}\right| + \left|\hat{y}_{i,t}\right|}\right)}{n} \times 100$$

The NSE is calculated as:

$$NSE = 1 - \frac{\sum (y_{i,t} - \hat{y}_{i,t})^2}{\sum (y_{i,t} - \bar{y}_{i,t})^2}$$

where $y_{i,t}$ is the observed rainfall at station $i$ and time $t$, $\hat{y}_{i,t}$ is the predicted rainfall, $\bar{y}_{i,t}$ is the mean observed rainfall at station $i$ over time, and $n$ is the total number of stations and time steps. NSE $< 0$ indicates that the predictive skill is worse than using the mean observed rainfall.

## 6 Empirical Results

Based on the dataset and the experimental setup in Section 5, we report the graph-based model results and attention weights for each cluster in Subsection 6.1. We then present examples of statistics of Spatial Season-aware GPD mapping in Subsection 6.2 and compare predictive skill with related work in Subsection 6.3. Finally, Subsection 6.4 presents operational forecasts evaluated against the ECMWF SEAS5 seasonal forecasts.

### 6.1 GNN Model Interpretability for Each Cluster of Stations

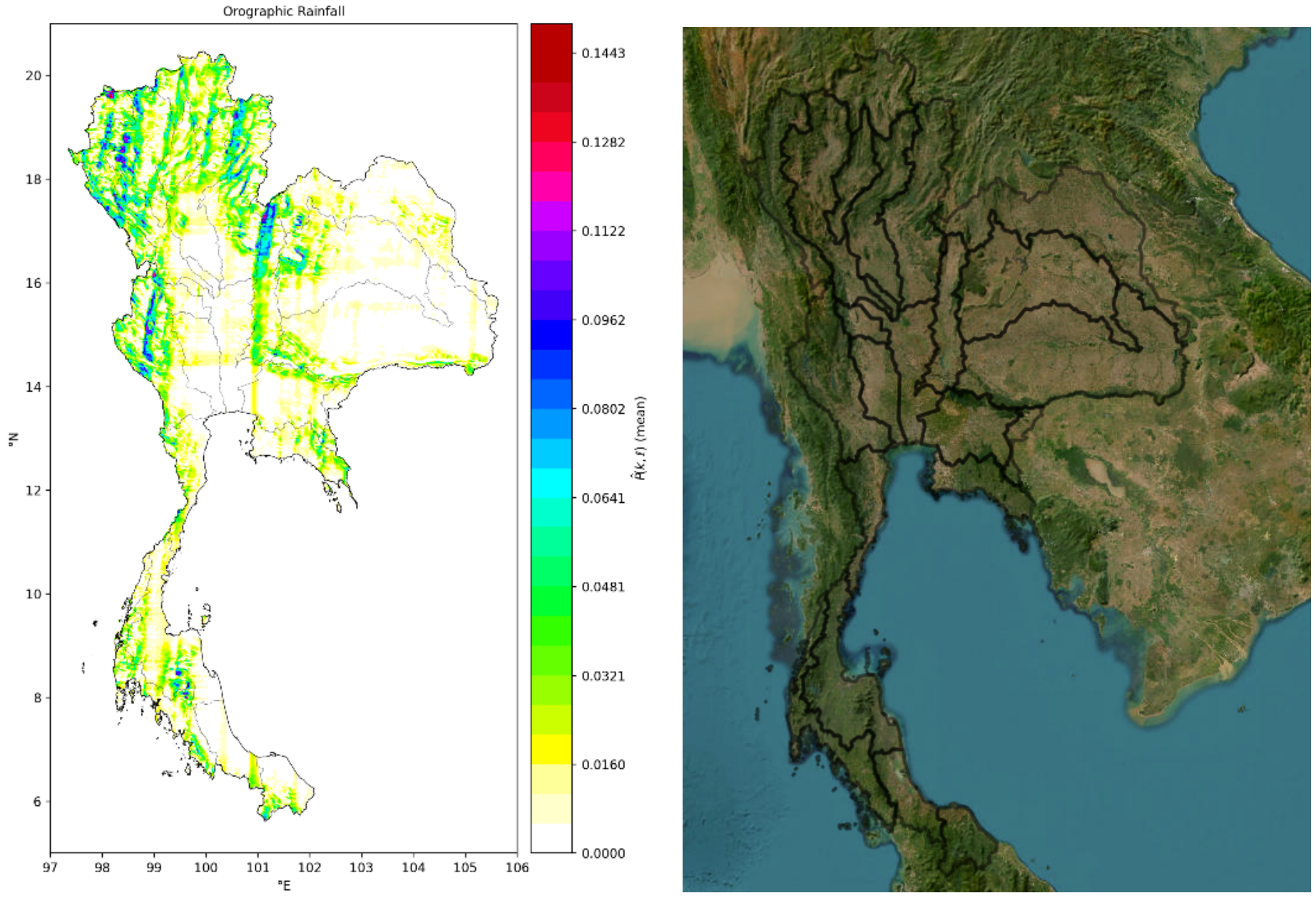

(a) The spectral transform of the precipitation $\hat{P}(k, \mathrm{l})$, simulated using ERA5 data spanning 1982–2024, compared with orography and the land-sea boundary in Thailand.

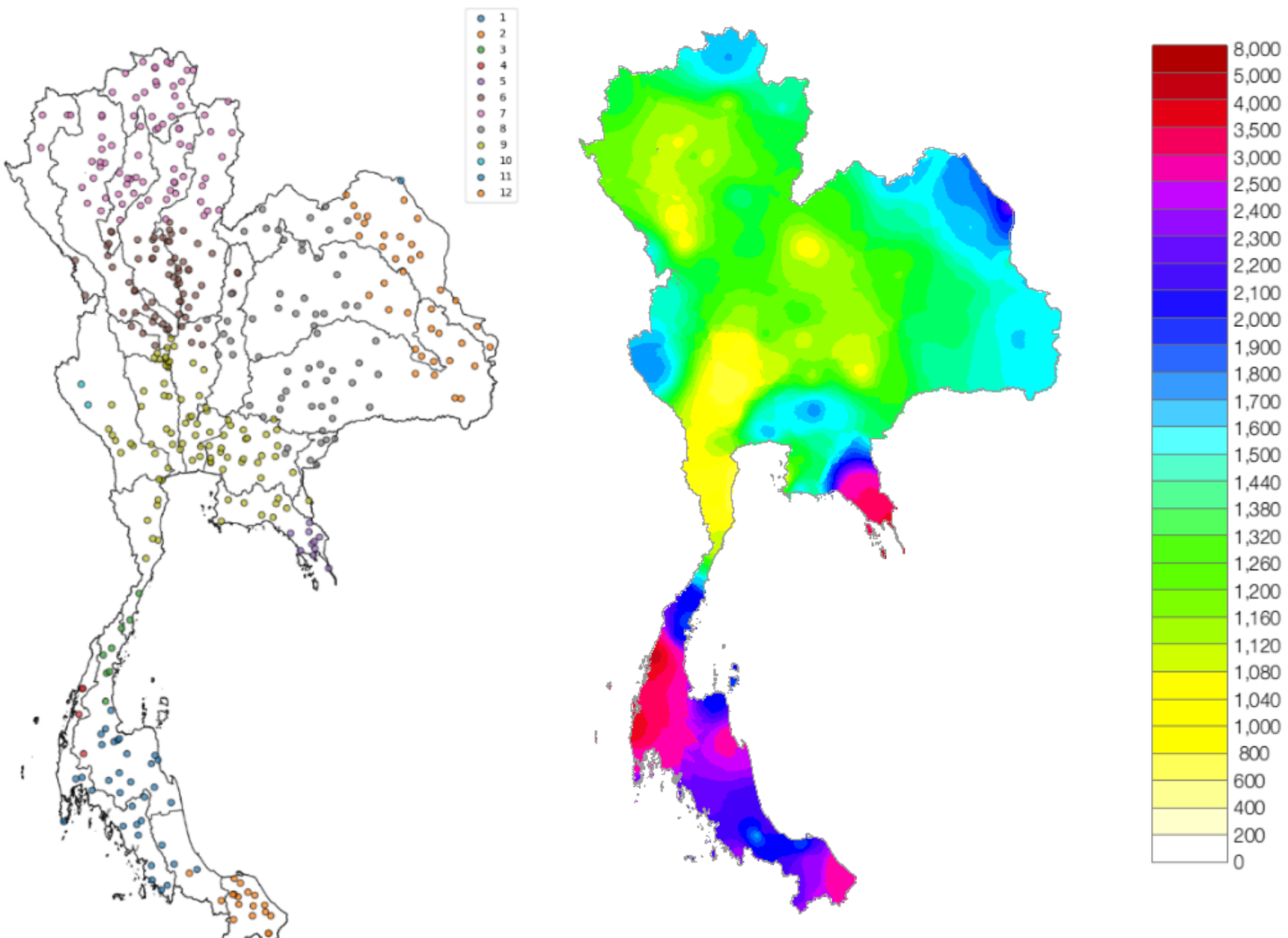

(b) Twelve-cluster results for HII rainfall stations and river basins, obtained by hierarchical clustering of HII rainfall stations (Algorithm 1), compared with the 30-year average TMD rainfall in Thailand (1991-2020).

**Figure 4.** Reanalysis graph-based structure (a) the spectral transform of the precipitation $\hat{P}(k, \mathrm{l})$ and (b) twelve-cluster areas for HII rainfall stations

The spectral transform of precipitation $\hat{P}(k, l)$ for each TMD rainfall station, simulated using monthly average ERA5 U and V wind components for 1982–2024, is shown in Figure 4a. A clear relationship between precipitation and high elevation is apparent. Monthly rainfall at each HII station serves as input to the hierarchical clustering algorithm. The results show that the 12-cluster areas indicate the spatial pattern of average monthly rainfall and geography across Thailand (Figures 4a and 4b). For example, in Figure 4b, Clusters 1 and 3 refer to station areas in Southern Thailand facing the Gulf of Thailand, while Cluster 4 represents areas along the Indian Ocean. Cluster 2 lies in the area closest to the ITCZ. Cluster 11 contains only one station in Nong Khai province, which frequently lies along storm paths. However, when we applied this clustering area to TMD stations, the sample size was too small to be used for training GAT-detrend.

For each cluster, we construct station-level graph structures using Pearson correlation and time-lag tests from monthly data. We define a teleconnection when the average absolute correlation during extreme events is higher than 0.4 and the time-lagged correlation is statistically significant (p-value < 0.1; see details in Appendix A1-A2). The differences in teleconnections between the models of Clusters 1 and 8 are BSISO and PDO. This aligns with the studies by Abatan et al. (2023) and Kikuchi (2021), which found that BSISO modulates sub-seasonal to seasonal rainfall in Northeast Thailand and is characterized by an elongated convective band orienting from the northwest to the southeast. All graph structures are reported in Appendix B.

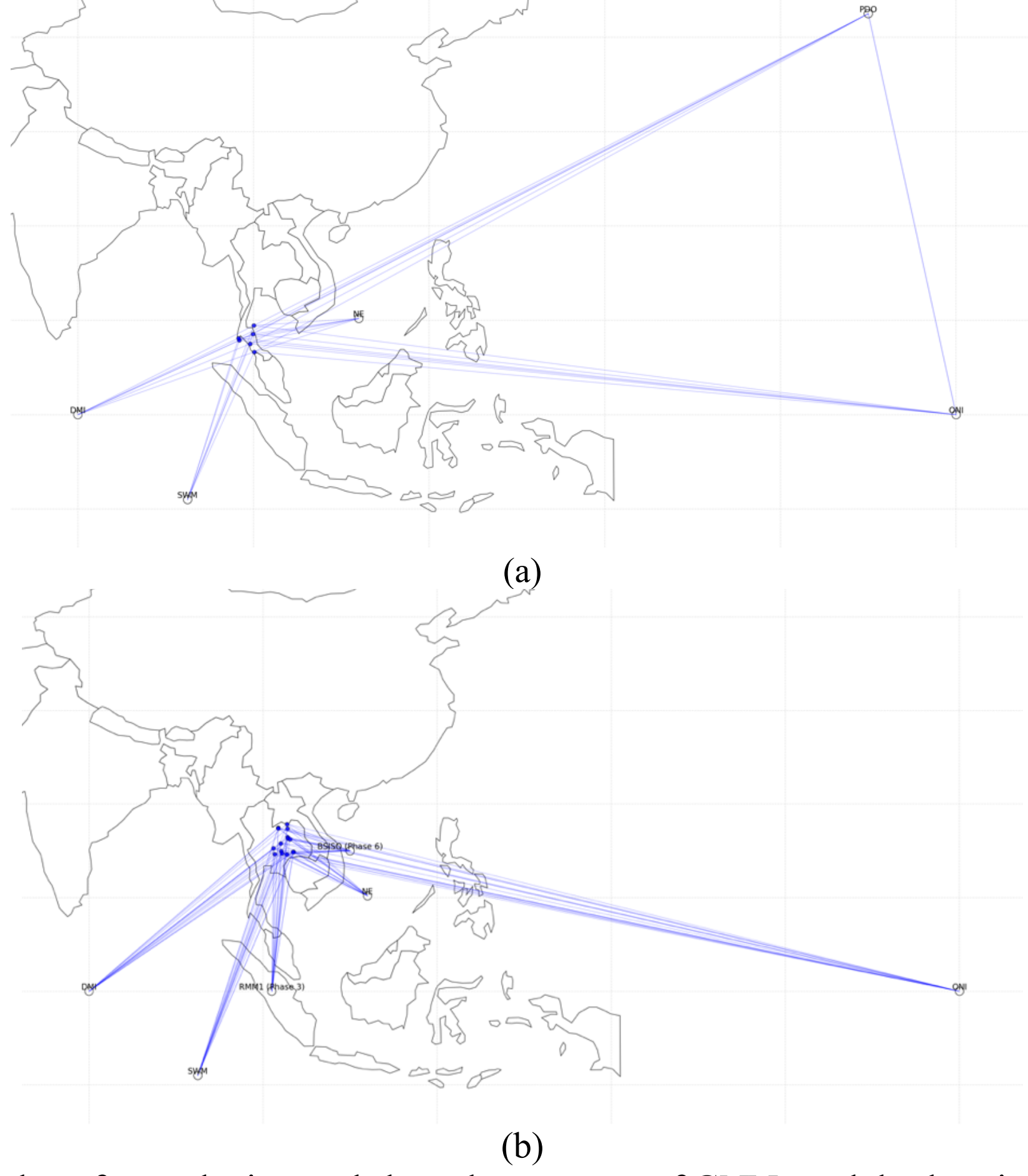

(a)

(b)

**Figure 5.** Examples of reanalysis graph-based structures of GNN models showing their stations as nodes and their teleconnections as edges in Cluster 1 (a) where DMI and PDO are lagged by 1 month and Cluster 8 (b).

### 6.1.1 Attention weights for each cluster

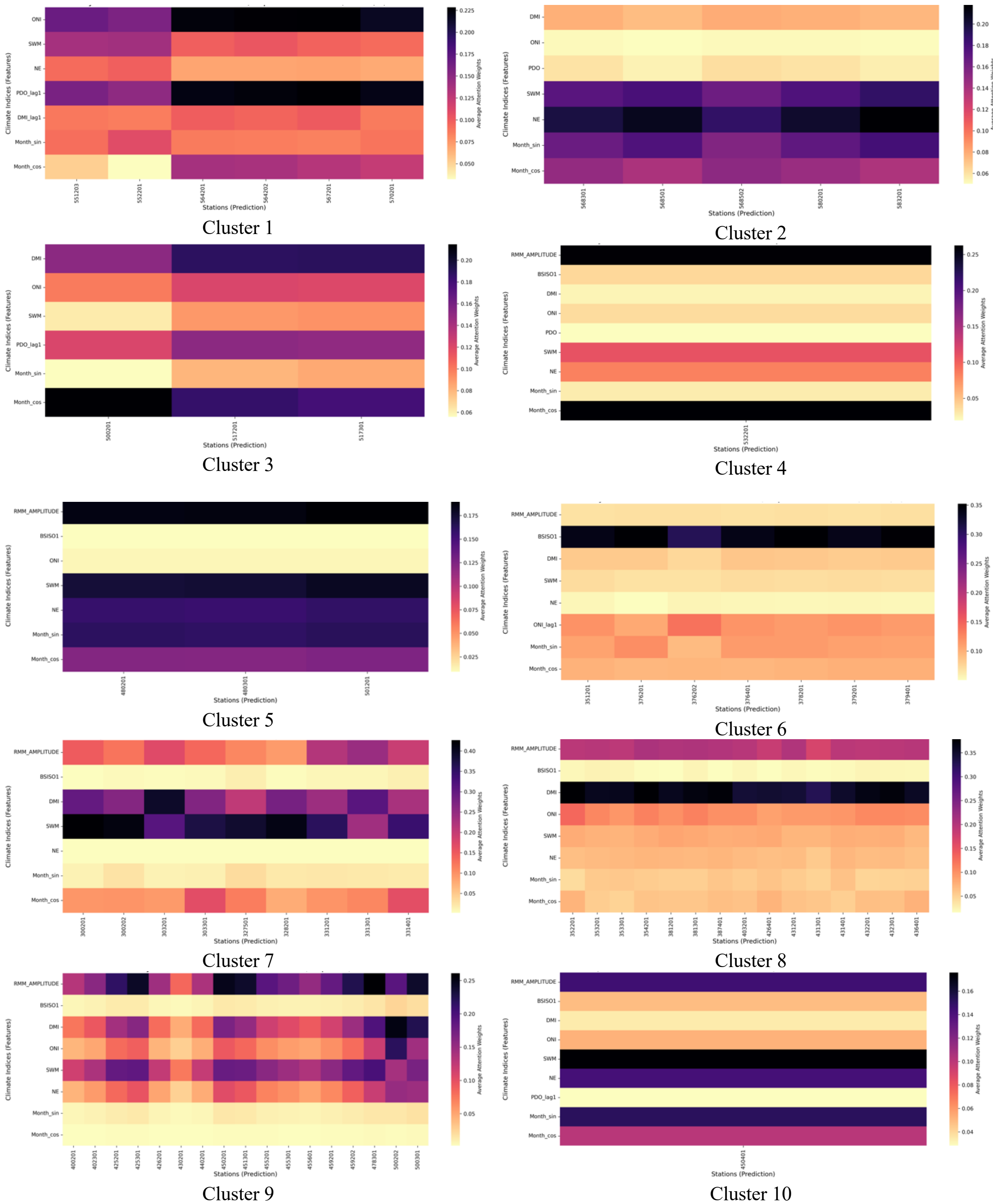

Cluster 1

Cluster 2

Cluster 3

Cluster 4

Cluster 5

Cluster 6

Cluster 7

Cluster 8

Cluster 9

Cluster 10

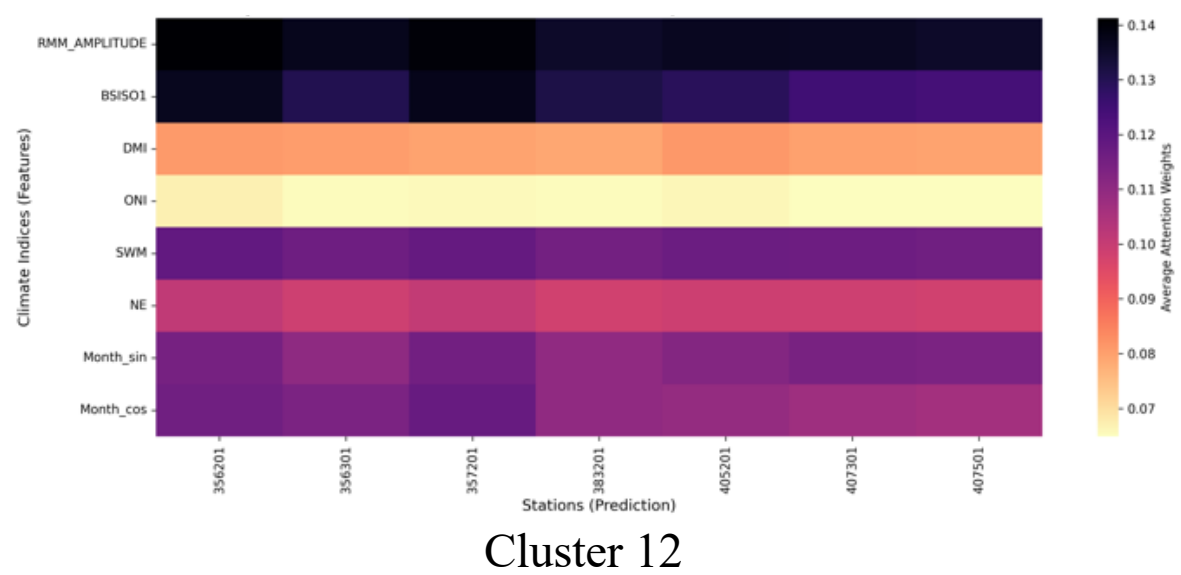

Cluster 12

**Figure 6.** The heatmaps showing average attention weights assigned by GAT-detrend to each climate index feature across all stations for Clusters 1–12. Darker colors indicate higher attention weight, reflecting stronger teleconnection of climate drivers influence on rainfall prediction.

To demonstrate the interpretability of GAT-detrend in this subsection, the attention heatmaps in Figure 6 demonstrate that the model captures physically meaningful patterns, as evidenced by the average attention weights. In the Southern region, Cluster 1 is primarily driven by SWM, PDO, and ONI, while Cluster 2 is influenced by both the SWM and NEM. Clusters 3 and 9 exhibit balanced attention across multiple climate indices, indicating complex, multi-driver transitions. Eastern Thailand (Cluster 5) is heavily influenced by the RMM (MJO), SWM, and NEM. Northern Thailand (Cluster 7) is dominated by the SWM and the DMI. In Northeastern Thailand, Cluster 12 is dominated by RMM (MJO) and BSISO. This aligns with Abatan et al. (2023) and shows balanced influences from various monsoon transitions and intraseasonal climate indices. By prioritizing these large-scale climate indices over simple temporal seasonality, the model provides transparent and interpretable signal attribution across Thailand's diverse rainfall regimes.

### 6.2 Results of Spatial Season-aware GPD for TMD stations

We perform Spatial Season-aware GPD mapping (Subsection 4.1) to analyse extreme values. Table 4 reports the examples of GPD parameters, estimated from MLE for examples of stations and seasons in Clusters 1 and 8, using the threshold at the 95$^{th}$ percentile (Q = 95). Exceedances refer to the number of observations above Q (i.e., extreme rainfall events). Because the same percentile Q is applied to training time series of each station, the number of exceedances for each season is the same across stations.

For station 567201 in Cluster 1, the observed threshold $u$ in the NE monsoon is 345.8 mm/month and in the SW monsoon is 362.7 mm/month. The predicted threshold $u$ computed from the fine-tuned GAT-detrend (see details in Table 6), given training data (Fold = 2), is 297.6 mm ($Q_{i\,,NE}(y)$) in the NE monsoon and 282.8 mm/month ($Q_{i\,,SW}(y)$) in the SW monsoon. For station 387401 in Cluster 8, the observed $u$ is 283.8 mm ($Q_{i\,,peak}(y)$) during the monsoon peak, and the predicted $u$ is 180.66 mm/month ($Q_{i\,,peak}(y)$) during the monsoon peak. These results indicate that the ML-predicted values of GAT-detrend are typically lower than the observed extreme rainfall. The charts of CDFs of GPD are shown in Figure 7. These patterns indicate a consistent underestimation of the tail of distributions by predicted values of GAT-detrend, most pronounced during the NE-monsoon months of Cluster 1 and during the monsoon-peak months of Cluster 8.

Consequently, Spatial Season-aware GPD mapping is particularly valuable for adjusting predicted extremes upward during the monsoon and monsoon-onset seasons. This offers a smooth mapping function for long-term rainfall forecasting. However, ($Q_{i,dry}(y)$) should not be applied during the dry season in these clusters.

| Node | Season | Observation | | | | Prediction of GAT-detrend | | | |
|---|---|---|---|---|---|---|---|---|---|
| | | $u$ | $\xi$ | $a_u$ | exceedances | $u$ | $\xi$ | $a_u$ | exceedances |
| Cluster 1 | | | | | | | | | |
| 567201 | Dry | 164.9 | 0.1 | 64.1 | 26 | 170.7 | -0.7 | 35.8 | 23 |
| | NE monsoon | 345.8 | -0.2 | 102.3 | 20 | 297.6 | -0.6 | 48.9 | 17 |
| | SW monsoon | 362.7 | -0.1 | 115.7 | 33 | 282.8 | -0.4 | 40.4 | 29 |
| Cluster 8 | | | | | | | | | |
| 387401 | Dry | 23.07 | 0.24 | 14.21 | 26 | 76.63 | -0.24 | 16.20 | 23 |
| | Monsoon onset | 188.32 | -0.13 | 80.22 | 20 | 124.37 | -0.44 | 19.36 | 17 |
| | Monsoon peak | 283.79 | -0.24 | 98.03 | 33 | 180.66 | -0.45 | 24.15 | 29 |

**Table 4.** The GPD fitted parameters obtained using MLE for the stations and their season in Clusters 1 and 8 where percentile Q is 95.

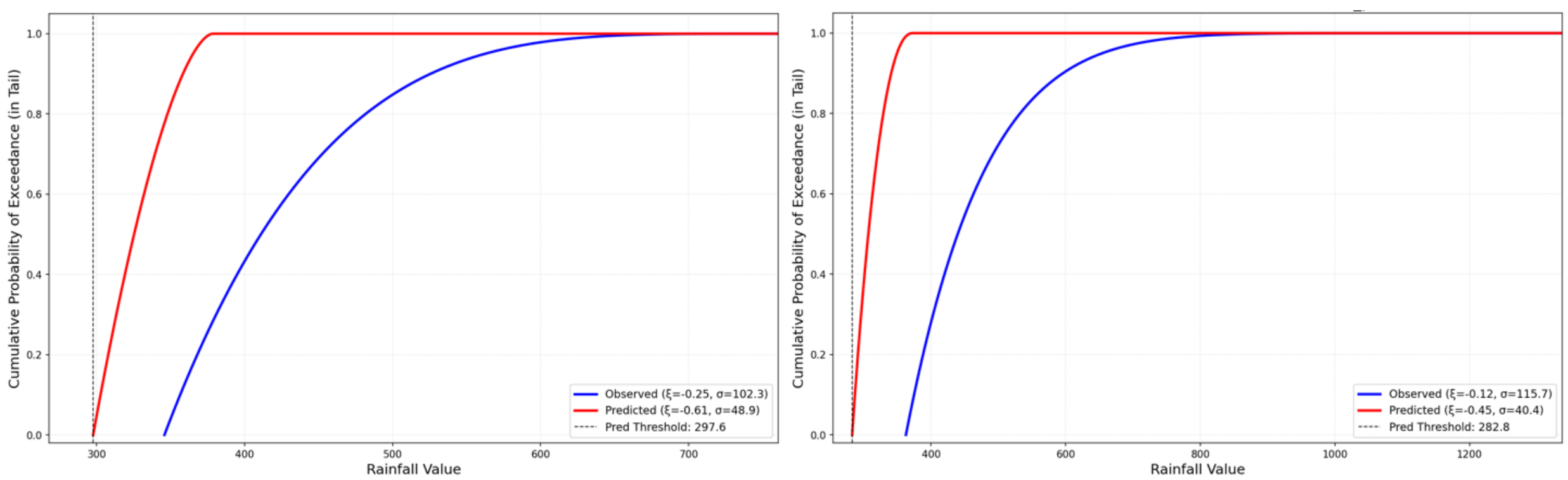

(a) CDFs for station 567201 in the NE monsoon season (b) CDFs for station 567201 in the SW monsoon

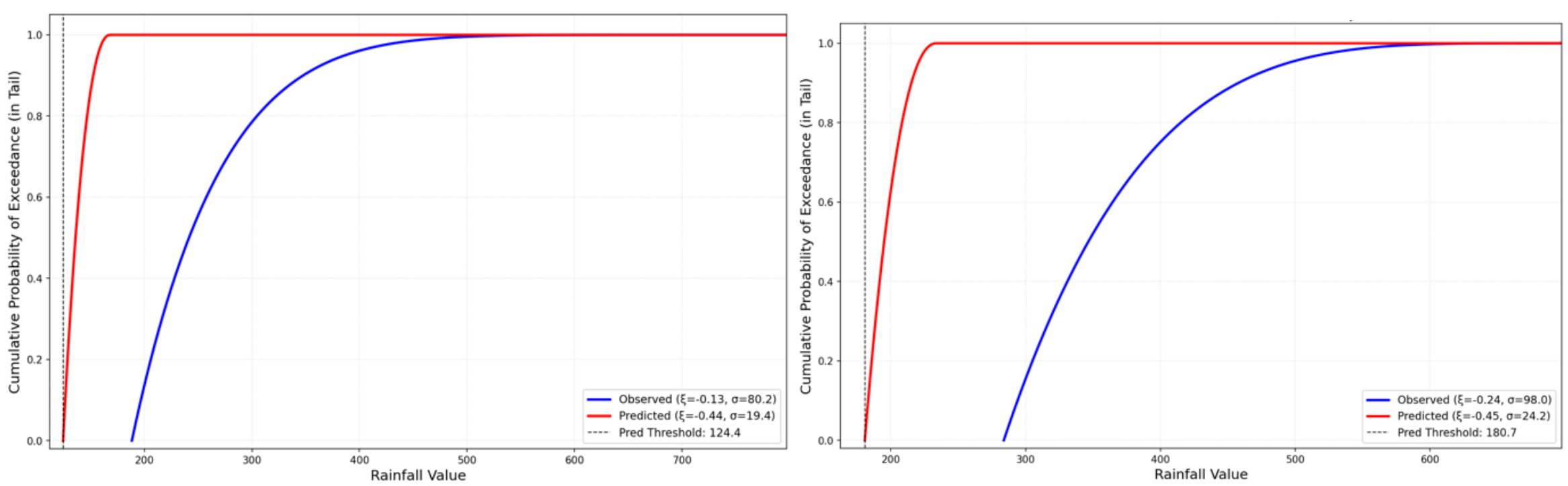

(c) CDFs for station 387401 in the monsoon onset (d) CDFs for station 387401 in the monsoon peak season

**Figure 7.** Example plots of CDFs of GPD for sample stations and their wet seasons where percentile is 95

### 6.3 Predictive Performance Metrics for TMD Stations

Table 5 details the best performance achieved by each algorithm and the eleven meteorological clusters of TMD stations when tuning the hyperparameters for specific metrics (accuracy, RMSE or NSE). The optimal dropout, number of layers, hidden layer size, and multi-head settings when using optimising accuracy are shown in Table 6. The results illustrate averages over stations within each cluster and forecast horizons. We evaluated GAT-detrend (without physics and GPD) and the GAT-detrend with GPD mapping at only exceedance cutoff percentile Q = 95. Note that the cutoff percentiles at Q = 90 and Q = 95 provide similar results, we then use the percentile Q = 95 throughout this study. We assessed the performance against baseline LSTM, Wavelet-LSTM, GC-LSTM, and EvolveGCN models. Note that Spatial Season-aware GPD mapping does not improve the performance of the competing approaches, and hence the results are not shown in the table. Overall, GAT-detrend achieves the highest average accuracy across all clusters. GAT-detrend mostly outperforms the competing algorithms in terms of accuracy, RMSE, and NSE. This indicates that GAT-detrend with physics and the Spatial Season-aware GPD mapping tends to yield higher performance metrics in most clusters among the competing approaches: LSTM, Wavelet-LSTM, GC-LSTM, and EvolveGCN.

Specifically, for averaged accuracy over Clusters 1–4, where the regimes experience more extreme rainfall, the maximum average accuracy of tuned models of GAT-detrend (Q = 95) is 70.56% compared with 62.05% for GAT-detrend (without physics and GPD), 60.96% for LSTM, 43.34% for Wavelet-LSTM, 57.12% for GC-LSTM, and 52.14% for EvolveGCN. The empirical results show that the attention mechanism, the physics of orographic precipitation, and Spatial Season-aware GPD mapping effectively improve predictive skill compared to other models. This is because GAT-detrend can dynamically update the attention weights for teleconnections between ONI, DMI, monsoon indices (SWM, NEM), and intraseasonal variability (BSISO, MJO). However, the averaged accuracy results over Clusters 5–12 show that GAT-detrend and LSTM perform comparably (68.87% and 66.49% respectively), reflecting heterogeneity in teleconnection strength across different regions, as the regimes of Clusters 5–12 experience fewer extreme rainfall events than those of Clusters 1–4.

Averaged across clusters, GAT-detrend achieves the highest mean accuracy (70%) compared with LSTM (64%), indicating consistent benefits from incorporating the physics of orographic precipitation and extreme-value analysis with the Spatial Season-aware GPD tail mapping. Mapping the tail of distributions at Q = 95 improves accuracy by 8% in relative terms over GAT-detrend (without physics and GPD). Relative to competing models in the southern Clusters 1–4, GAT-detrend outperforms LSTM by 16%, Wavelet-LSTM by 63%, GC-LSTM by 24%, and EvolveGCN by 35% in relative terms. These gains indicate that tail-focused processing enhances predictive skill, consistent with the importance of rainfall extremes. Moreover, the stronger threshold (Q = 95) yields the most pronounced advantage in this region.

The physics of orographic precipitation and the Spatial Season-aware GPD tail mapping markedly lower the average RMSE in several clusters: for example, with GAT-detrend, RMSE decreases from 126.69, 89.80, 136.72, and 130.82 mm/month to 90.45, 74.68, 103.6, and 113.57 mm/month of GAT-detrend (without physics and GPD) in Clusters 1-4 respectively. Compared with the LSTM baseline, RMSE with GAT-detrend is 90.45 versus 133.65 for LSTM in Cluster 1; 74.68 versus 91.59 in Cluster 2; 103.6 versus 145.36 in Cluster 3; 113.57 versus 133.79 in Cluster 4. GAT-detrend achieves the highest average NSE in all clusters (e.g., Cluster 1: 0.72; Cluster 4: 0.86; Cluster 5: 0.82; Cluster 7: 0.81; Cluster 10: 0.81), indicating improved correspondence between predicted and observed rainfall compared with using simple averages of observed rainfall values.

In terms of computational time, models use different hyperparameter sets with early stopping enabled, and hence the training time for each hyperparameter combination and total run time is not directly comparable. The average training time per epoch for GAT-detrend, Wavelet-LSTM, GC-LSTM, LSTM, and EvolveGCN is 1.24 sec, 1.53 sec, 2.39 sec, 1.06 sec, and 1.17 sec respectively.

| Cluster (stations) | GAT-detrend (without physics and GPD) | GAT-detrend (Q = 95) | LSTM | Wavelet-LSTM | GC-LSTM | Evolve-GCN |
|---|---|---|---|---|---|---|
| | **Maximum average accuracy** | | | | | |
| 1 (TMD: 6) | 64.32 | **67.93** | 58.50 | 39.48 | 59.34 | 52.52 |
| 2 (TMD: 5) | 63.36 | **68.41** | 61.65 | 42.62 | 61.50 | 53.82 |
| 3 (TMD: 3) | 53.92 | **64.82** | 53.61 | 42.54 | 53.56 | 49.33 |
| 4 (TMD: 1) | 66.60 | **81.08** | 70.08 | 48.72 | 54.09 | 52.87 |
| 5 (TMD: 3) | 64.68 | **68.76** | 67.20 | 42.27 | 42.58 | 60.47 |
| 6 (TMD: 7) | 58.63 | **65.38** | 65.00 | 55.40 | 34.82 | 58.71 |
| 7 (TMD: 9) | 71.14 | **73.65** | 66.31 | 55.94 | 41.23 | 66.80 |
| 8 (TMD: 15) | 64.60 | 65.65 | **66.04** | 61.98 | 34.52 | 63.32 |
| 9 (TMD: 17) | 62.62 | 64.03 | 62.07 | **64.45** | 37.48 | 60.69 |
| 10 (TMD: 1) | 67.49 | **73.26** | 70.54 | 38.43 | 39.67 | 65.65 |
| 11 (TMD: 0) | - | - | - | - | - | - |
| 12 (TMD: 7) | 70.05 | **71.37** | 68.30 | 66.73 | 42.11 | 66.63 |
| | **Minimum average RMSE (mm/month)** | | | | | |
| 1 (TMD: 6) | 126.69 | **90.45** | 133.65 | 179.11 | 126.90 | 160.90 |
| 2 (TMD: 5) | 89.80 | **74.68** | 91.59 | 140.70 | 86.96 | 128.56 |
| 3 (TMD: 3) | 136.72 | **103.60** | 145.36 | 173.32 | 151.41 | 157.62 |
| 4 (TMD: 1) | 130.82 | **113.57** | 133.79 | 142.15 | 133.96 | 186.85 |
| 5 (TMD: 3) | 192.40 | **157.62** | 230.11 | 240.46 | 200.37 | 206.01 |
| 6 (TMD: 7) | 62.57 | **58.01** | 61.16 | 62.32 | 59.59 | 68.28 |
| 7 (TMD: 9) | 56.22 | **54.21** | 55.52 | 65.30 | 55.37 | 62.93 |
| 8 (TMD: 15) | 72.23 | **64.74** | 69.46 | 72.32 | 69.79 | 75.50 |
| 9 (TMD: 17) | 59.47 | **58.40** | 64.53 | 60.97 | 59.67 | 67.85 |
| 10 (TMD: 1) | 102.90 | **76.57** | 99.11 | 158.74 | 100.23 | 118.27 |
| 11 (TMD: 0) | - | - | - | - | - | - |
| 12 (TMD: 7) | 65.67 | 62.56 | 64.36 | 71.86 | **59.91** | 76.61 |
| | **Maximum average NSE** | | | | | |
| 1 (TMD: 6) | 0.61 | **0.72** | 0.50 | -0.01 | 0.56 | 0.34 |
| 2 (TMD: 5) | 0.60 | **0.70** | 0.60 | 0.04 | 0.63 | 0.48 |
| 3 (TMD: 3) | 0.44 | **0.61** | 0.35 | -0.01 | 0.29 | 0.23 |
| 4 (TMD: 1) | 0.75 | **0.86** | 0.69 | 0.00 | 0.68 | 0.39 |
| 5 (TMD: 3) | 0.67 | **0.82** | 0.60 | -0.03 | 0.66 | 0.57 |
| 6 (TMD: 7) | 0.59 | **0.67** | 0.64 | 0.61 | 0.65 | 0.55 |
| 7 (TMD: 9) | 0.73 | **0.81** | 0.68 | 0.47 | 0.72 | 0.58 |
| 8 (TMD: 15) | 0.58 | **0.65** | 0.61 | 0.58 | 0.61 | 0.55 |
| 9 (TMD: 17) | 0.64 | **0.65** | 0.59 | 0.63 | 0.64 | 0.54 |
| 10 (TMD: 1) | 0.65 | **0.81** | 0.63 | 0.06 | 0.65 | 0.48 |
| 11 (TMD: 0) | - | - | - | - | - | - |
| 12 (TMD: 7) | 0.69 | **0.75** | 0.70 | 0.67 | **0.75** | 0.65 |

**Table 5.** The maximum average accuracy, the minimum average RMSE, and the maximum average NSE accuracy across all horizons and TMD stations for each of the six tuned models for each hyperparameter setting. The best performance metrics are shown in bold.

Therefore, the explicit tail-mapping method, Spatial Season-aware GPD, is effectively combined with the attention mechanism of GAT-detrend and physics. It generally yields the highest accuracy and NSE. However, the reductions in error magnitude occur primarily in southern Thailand (Clusters 1–4), where heavy-rainfall regimes make tail mapping especially beneficial. However, not all clusters benefit equally. These findings support the effectiveness of physics-informed GAT-detrend paired with Spatial Season-aware GPD mapping for long-range monthly rainfall prediction in Thailand. In a few cases, LSTM and Wavelet-LSTM remain superior for accuracy, highlighting discrepancies between these metrics and accuracy. Moreover, both GC-LSTM and EvolveGCN use graph convolutional models whose Chebyshev polynomial spectral filters may not be appropriate for long-range rainfall prediction.

| Cluster | Multi head | Hidden Size | Number of layers | Dropout | Concat |
|---|---|---|---|---|---|
| 1 | 8 | 64 | 1 | 0.4 | False |
| 2 | 8 | 16 | 2 | 0.4 | True |
| 3 | 8 | 128 | 2 | 0.4 | True |
| 4 | 1 | 128 | 2 | 0.4 | True |
| 5 | 8 | 16 | 2 | 0 | False |
| 6 | 8 | 128 | 2 | 0.4 | True |
| 7 | 1 | 128 | 1 | 0 | True |
| 8 | 8 | 128 | 1 | 0.2 | False |
| 9 | 1 | 64 | 2 | 0.2 | True |
| 10 | 8 | 16 | 2 | 0.4 | True |
| 12 | 8 | 16 | 1 | 0.2 | True |

**Table 6.** Optimal hyperparameters of the fine-tuned GAT-detrend model using accuracy for each cluster of TMD stations.

### 6.4 Real-world Applications compared with ECMWF Seasonal Forecasts

In ECMWF operational seasonal forecasting system from SEAS5, probabilities of precipitation anomalies or the mean of precipitation over 51 ensembles summarise the expected climate over the coming months from a statistical view of potential weather. The operational web-based products are available here.
(https://charts.ecmwf.int/products/seasonal_system5_standard_rain?area=AUSG&stats=ensm)

To illustrate real-world applications in this study, we use out-of-sample data to predict rainfall of wet seasons from 74 ground-truth TMD stations for July–December 2025 in this subsection. As shown in Figure 3, inputs of the end-to-end workflow for long-range rainfall forecasts include gauge rainfall, reanalysis data (e.g., monthly OISST, U and V components of wind). The system processes climate indices (PDO, SWM, NEM, ONI, DMI, etc.) and computes initial edge features using Smith linear model. It generates monthly rainfall forecasts up to 12 months ahead. Model development and inference are conducted in version-controlled environments on HPC/GPU infrastructure, with fine-tuned hyperparameters to maximise the accuracy.

We generate high-resolution forecasting maps by interpolating TMD gauge observations using the Inverse Distance Weighting (IDW) method. Forecasts in Figures 8-13 are generated from ECMWF SEAS5 and from our operational long-range rainfall forecasting system (https://ai-long-range.hii.or.th/rain/forecast.html), which employs the GAT-detrend model. This is because the GAT-detrend model provides the highest average accuracy across clusters (see details in Table 5).

Both predictions are initialized in July 2025 with a six-month lead time, covering July through December 2025. Note that the inputs and outputs of ECMWF SEAS5 and our proposed model are different. We validate this comparison against 74 gauge stations used as sample grid points. In terms of monthly RMSE at these stations from July–December 2025, GAT-detrend achieves a better RMSE than SEAS5 by 21.83 mm/month on average and yields a better RMSE in about 14.6% of cases. SEAS5 slightly outperforms GAT-detrend in October. Considering average rainfall across Thailand, GAT-detrend accuracy is 79.3% compared to ground truth across six wet months on average whereas SEAS5 accuracy is 72.1% on average. Overall, these results indicate that GAT-detrend is suitable for operational downscaling for water-management applications, while the native SEAS5 resolution is comparatively coarse for community-level water management decision-making in real-world applications.

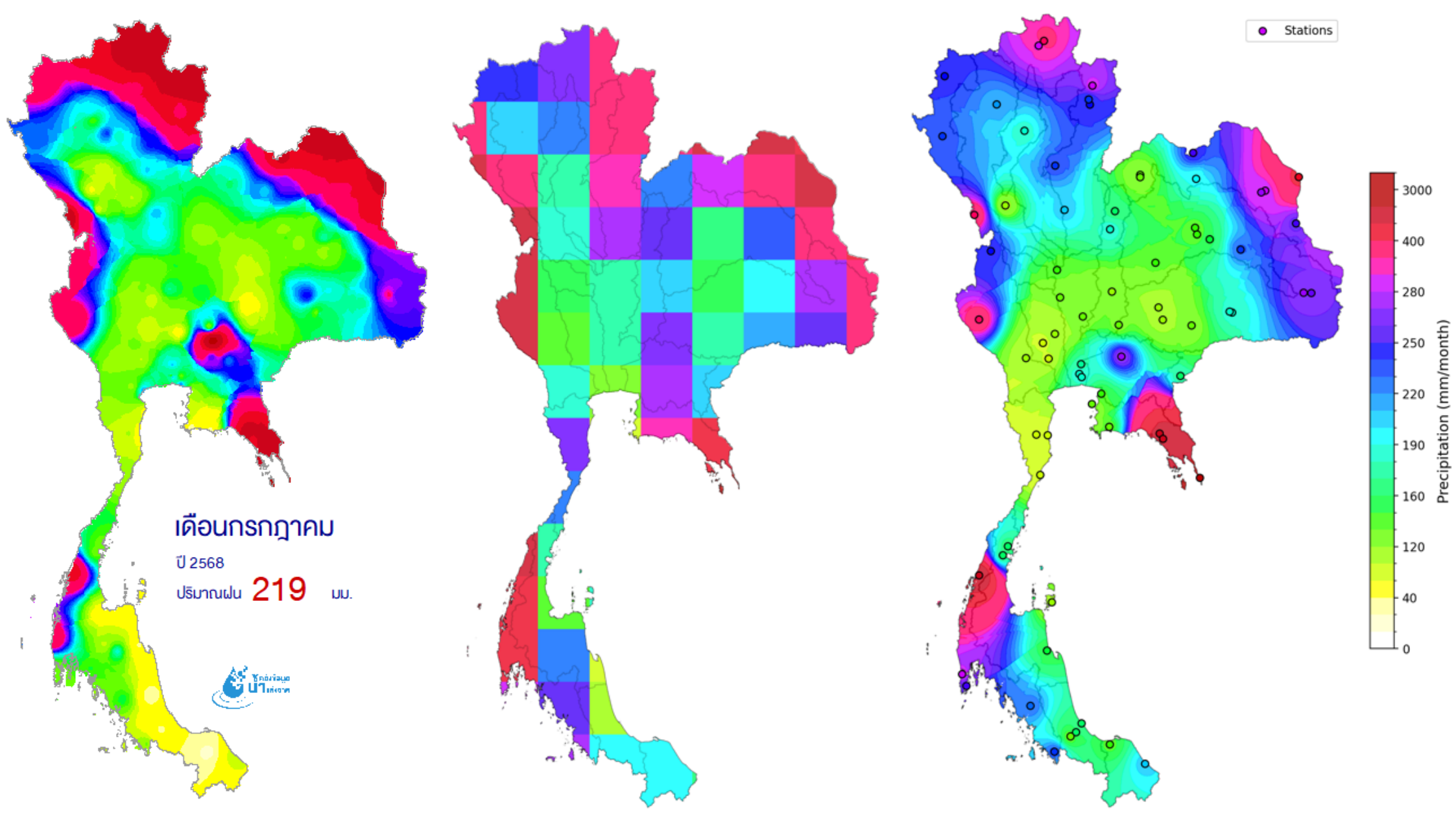

**Figure 8.** Monthly total precipitation for July 2025. Left: ground truth (average rainfall = 219 mm/month) Middle (prediction): SEAS5 (RMSE = 128 mm/month, average rainfall = 263 mm/month). Right (prediction): GAT-detrend (RMSE = 108 mm/month, average rainfall = 216 mm/month).

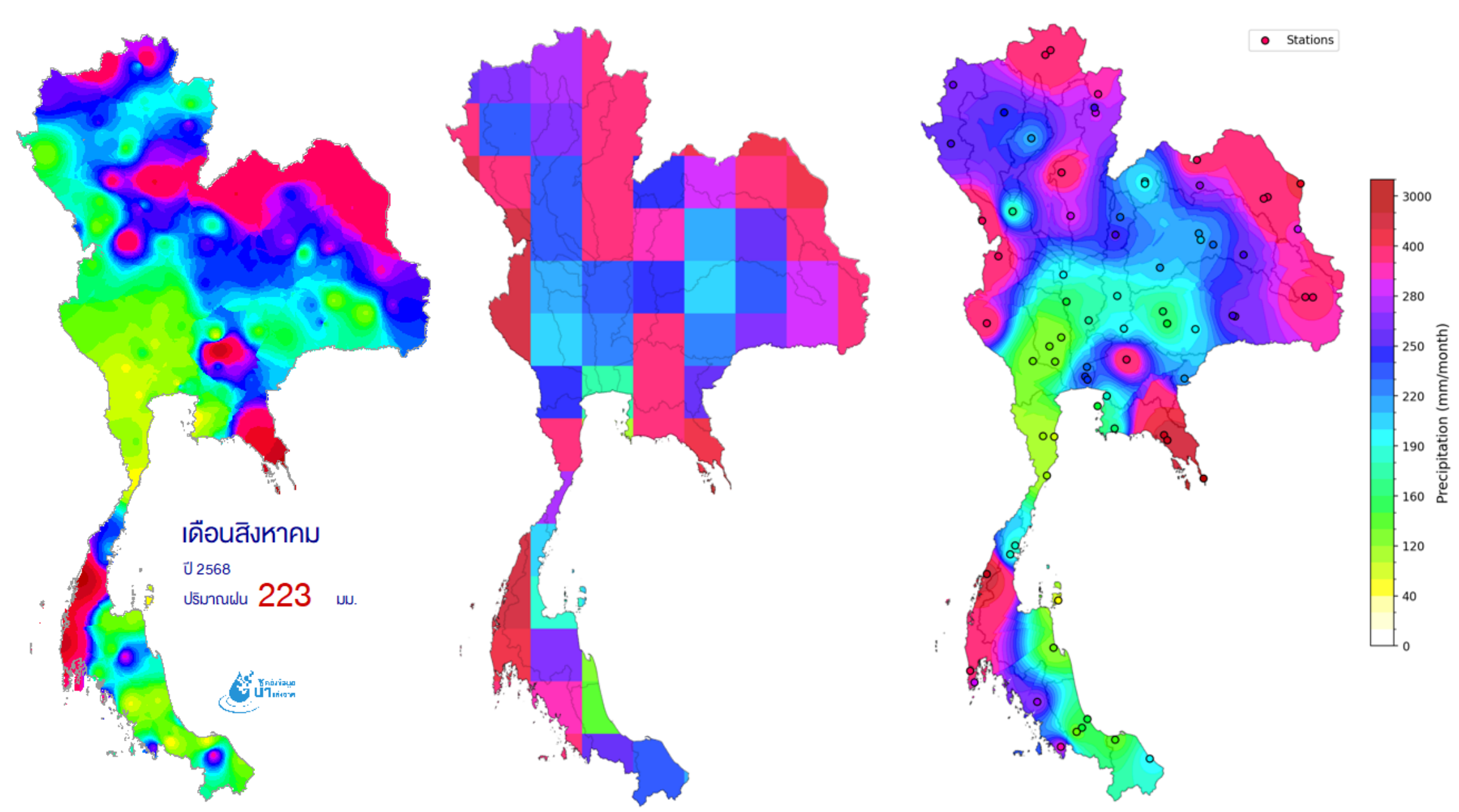

**Figure 9.** Monthly total precipitation for August 2025. Left: ground truth. (average rainfall = 223 mm/month) Middle (prediction): SEAS5 (RMSE = 161 mm/month, average rainfall = 293 mm/month). Right (prediction): GAT-detrend (RMSE = 105 mm/month, average rainfall = 248 mm/month).

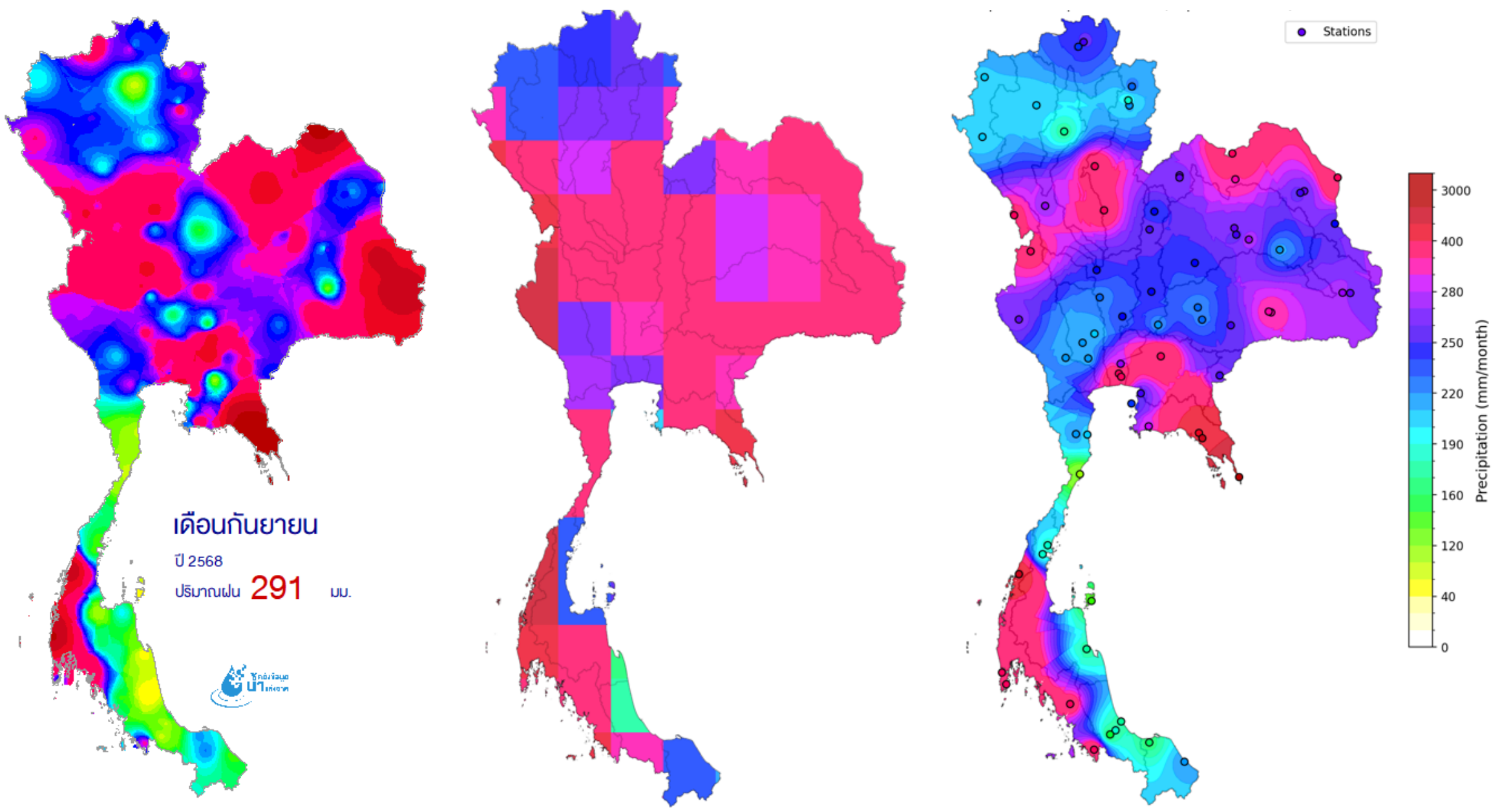

**Figure 10.** Monthly total precipitation for September 2025. Left: ground truth (average rainfall = 291 mm/month). Middle (prediction): SEAS5 (RMSE = 130 mm/month, average rainfall = 311 mm/month). Right (prediction): GAT-detrend (RMSE = 106 mm/month, average rainfall = 267 mm/month).

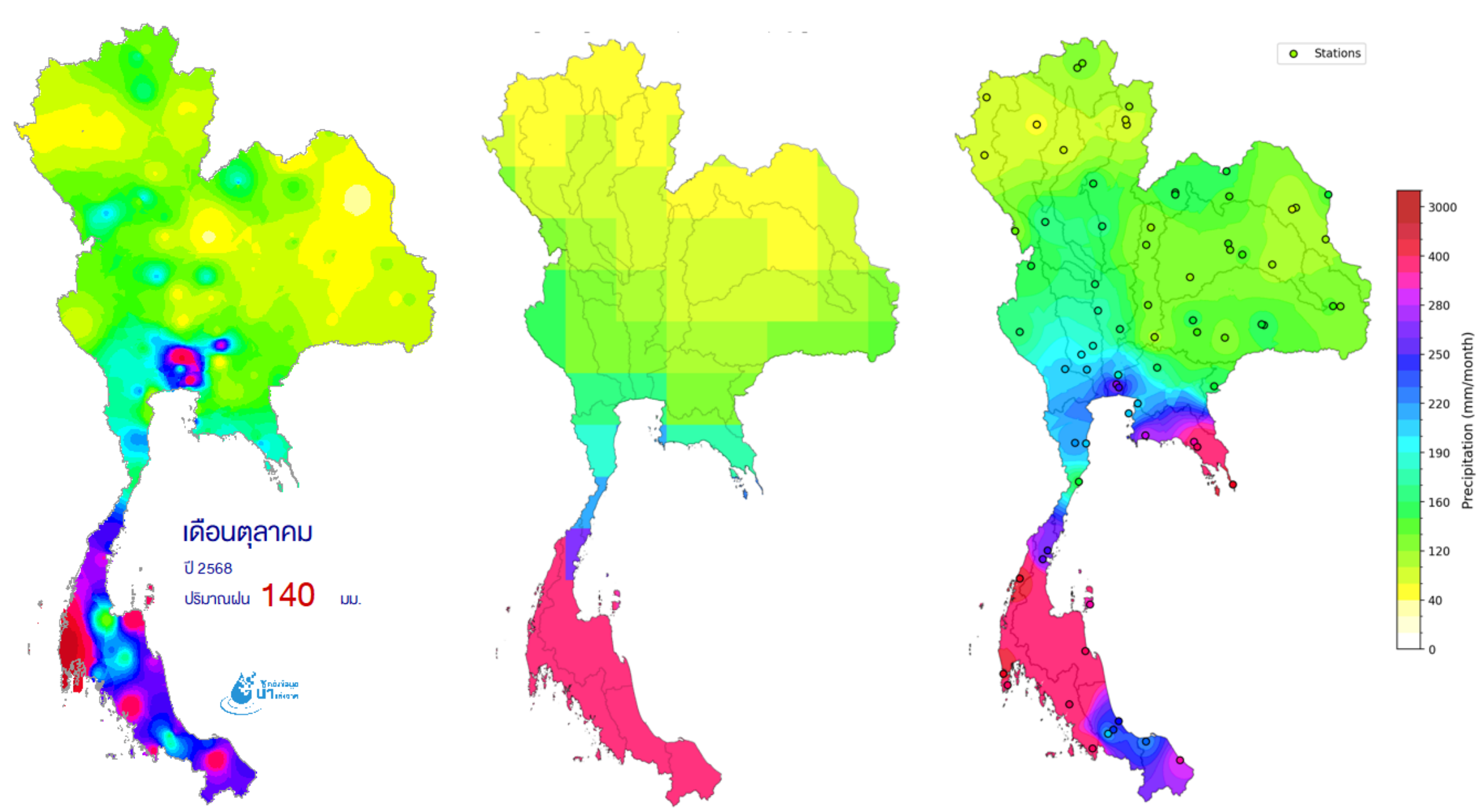

**Figure 11.** Monthly total precipitation for October 2025. Left: ground truth (average rainfall = 140 mm/month). Middle (prediction): SEAS5 (RMSE = 70 mm/month, average rainfall = 157 mm/month) Right (prediction): GAT-detrend (RMSE = 77 mm/month, average rainfall = 170 mm/month).

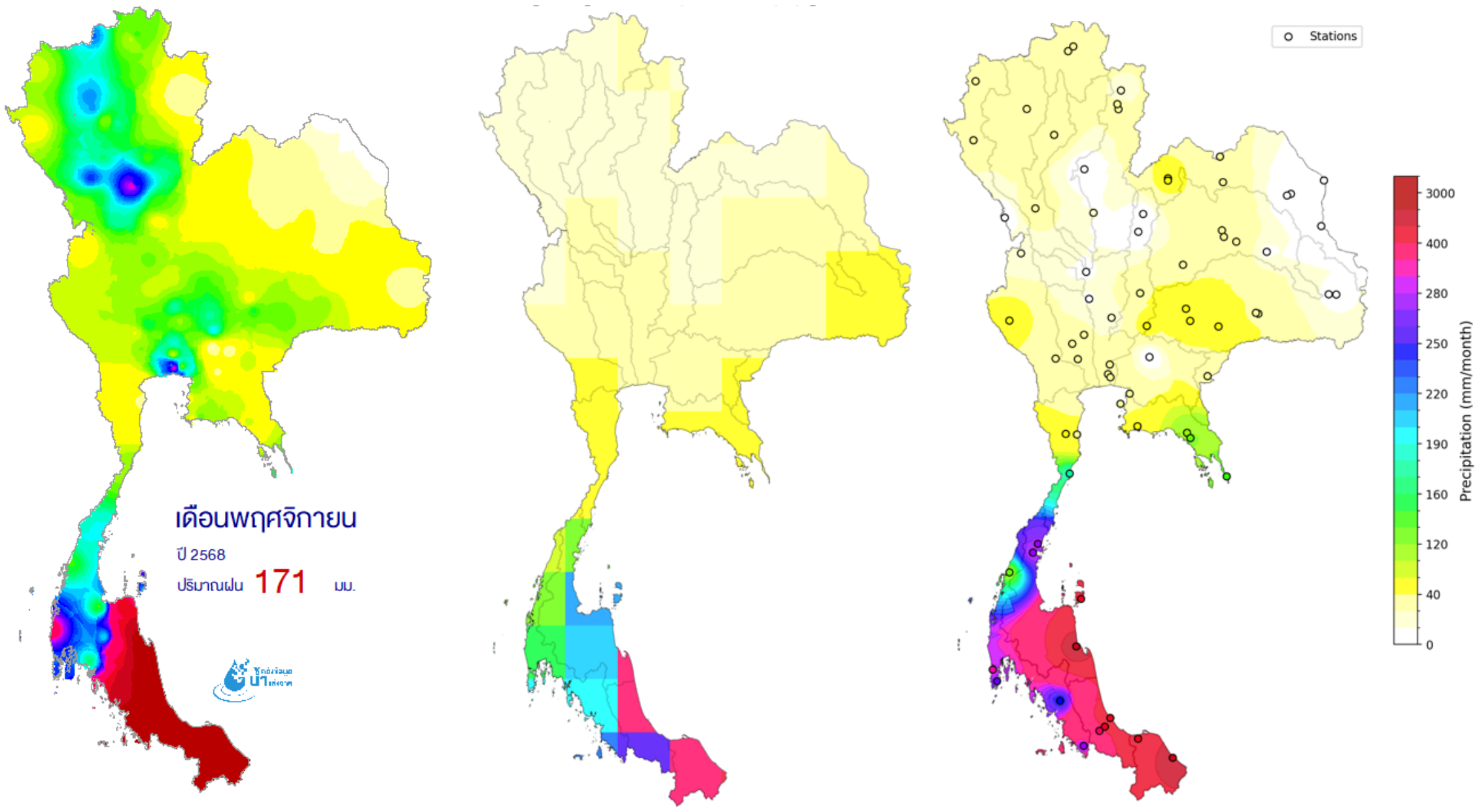

**Figure 12.** Monthly total precipitation for November 2025. Left: ground truth (average rainfall = 171 mm/month). Middle (prediction): SEAS5 (RMSE = 302 mm/month, average rainfall = 69 mm/month). Right (prediction): GAT-detrend (RMSE = 274 mm/month, average rainfall = 95 mm/month).

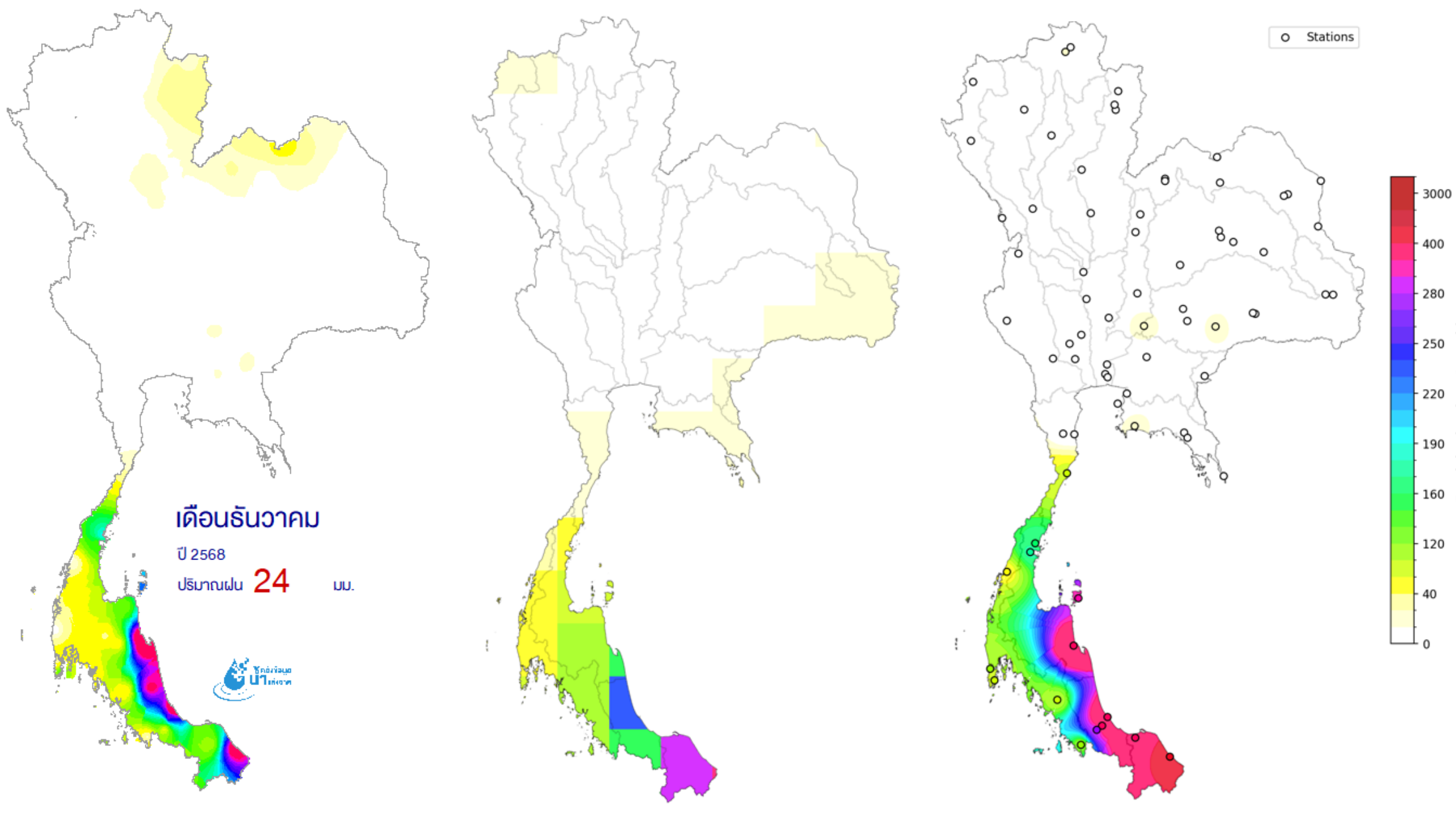

**Figure 13.** Monthly total precipitation for December 2025. Left: ground truth (average rainfall = 24 mm/month). Middle (prediction): SEAS5 (RMSE = 51 mm/month, average rainfall = 33 mm/month). Right (prediction): GAT-detrend (RMSE = 41 mm/month, average rainfall = 33 mm/month).

## 7 Conclusion and Future Work

This work presents a physics-informed graph-based model, GAT-detrend, for long-range rainfall forecasting at gauge stations, together with an operational forecasting system across Thailand. By combining physics-informed edge features, attention-based teleconnection learning, and a novel Spatial Season-aware GPD mapping to correct extreme-value underestimation in the distribution tails, the system produces forecasts 1–12 months ahead from historical rainfall using climate indices. Empirically, GAT-detrend outperforms established baselines across most clusters and achieves higher accuracy for extreme events. In a real-world comparison against ECMWF SEAS5 seasonal forecasts, GAT-detrend achieves a lower average RMSE, indicating practical value for operational, community-level water management.

Several enhancements remain for future work. First, replacing static graphs with learned or dynamic graphs could help capture evolving teleconnections. Second, augmenting spatial interpolation with ML-based methods such as DeepKriging (Chen et al., 2023) should be explored for high-resolution maps in real-world applications. Finally, combining the Spatial Season-aware GPD mapping with probabilistic calibration could further improve extreme-value performance for water-resource management.

Beyond Thailand, the proposed model is generalisable to other monsoon or topographically complex regions where teleconnection-driven rainfall variability limits conventional forecasting approaches. The graph-based model may also extend to other domains involving streamflow, or weather forecasting. The combination of interpretable attention weights and extreme-value analysis offers a practical tool for reservoir operations and agricultural planning. This emphasises the importance of Earth system science in the relationships among climate variables within explainable models and the correct implementation of ML methods rather than using them as black boxes.

**Acknowledgements**
The authors would like to express their gratitude to the Program Management Unit for Human Resources & Institutional Development, Research and Innovation (PMU-B), Ministry of Higher Education, Science, Research and Innovation, Thailand for funding this research.

**Declaration of generative AI and AI-assisted technologies in the manuscript preparation process**

During the preparation of this work the author(s) used Claude and Gemini to assist in implementing source code and editing the manuscript. After using this tool/service, the author(s) reviewed and edited the content as needed and take(s) full responsibility for the content of the published article.

## Appendix A.

**Table A1.** The maximum and minimum of average Pearson correlations (R) of climate index and rainfall across stations for each cluster between 2014-2024 used in experiments.

| Cluster | Climate Index | Max of R | Min of R |
|---|---|---|---|
| 1 | DMI | 0.64 | -0.54 |
| | NE | 0.54 | 0.54 |
| | ONI | 0.68 | -0.73 |
| | PDO | 0.62 | -0.48 |
| | RMM1 | -0.40 | -0.40 |
| | SWM | -0.41 | -0.60 |
| 2 | DMI | 0.64 | -0.55 |
| | ONI | 0.68 | -0.75 |
| | PDO | 0.72 | -0.58 |
| 3 | DMI | 0.62 | -0.68 |
| | NE | 0.52 | 0.52 |
| | ONI | 0.70 | -0.62 |
| | PDO | 0.55 | -0.69 |
| | RMM1 | -0.48 | -0.48 |
| | SWM | -0.40 | -0.65 |
| 4 | BSISO1 | 0.73 | 0.42 |
| | DMI | 0.70 | -0.81 |
| | NE | 0.67 | -0.45 |
| | ONI | 0.55 | -0.66 |
| | PDO | 0.57 | -0.53 |
| | RMM1 | -0.42 | -0.71 |
| | SWM | -0.67 | -0.83 |
| 5 | BSISO1 | 0.80 | 0.43 |
| | NE | 0.60 | -0.53 |
| | ONI | 0.65 | -0.53 |
| | RMM1 | -0.42 | -0.69 |
| | SWM | -0.54 | -0.80 |
| 6 | BSISO1 | 0.64 | 0.44 |
| | DMI | 0.54 | -0.66 |
| | NE | 0.49 | -0.49 |
| | ONI | 0.45 | -0.62 |
| | RMM1 | -0.40 | -0.56 |
| | SWM | -0.52 | -0.73 |
| 7 | BSISO1 | 0.66 | 0.40 |
| | DMI | 0.51 | -0.67 |
| | NE | 0.42 | -0.43 |
| | RMM1 | -0.47 | -0.52 |
| | SWM | -0.47 | -0.74 |
| 8 | BSISO1 | 0.64 | 0.47 |
| | DMI | 0.59 | -0.67 |
| | NE | 0.41 | -0.45 |
| | ONI | 0.42 | -0.49 |
| | RMM1 | -0.43 | -0.65 |
| | SWM | -0.43 | -0.77 |
| 9 | BSISO1 | 0.65 | 0.48 |
| | DMI | 0.62 | -0.65 |
| | ONI | 0.50 | -0.51 |
| | RMM1 | -0.47 | -0.62 |
| | SWM | -0.58 | -0.75 |
| 10 | BSISO1 | 0.61 | 0.50 |
| | DMI | 0.72 | -0.75 |
| | NE | 0.74 | -0.41 |
| | ONI | 0.55 | -0.64 |
| | PDO | 0.57 | 0.44 |
| | RMM1 | 0.48 | -0.66 |
| | SWM | -0.48 | -0.81 |
| 12 | BSISO1 | 0.64 | 0.43 |
| | DMI | 0.49 | -0.64 |
| | NE | 0.45 | -0.44 |
| | ONI | 0.53 | -0.52 |
| | RMM1 | -0.49 | -0.59 |
| | SWM | -0.49 | -0.68 |

**Table A2.** The average p-values given the Granger causality test of climate index (Y) for each cluster between 2014-2024 used in experiments.

| Cluster | Climate Index | Lag (month) | P-value |
|---|---|---|---|
| 1 | DMI | 1 | 0.082 |
| 1 | PDO | 1 | 0.056 |
| 3 | PDO | 1 | 0.073 |
| 10 | PDO | 1 | 0.064 |

## Appendix B.

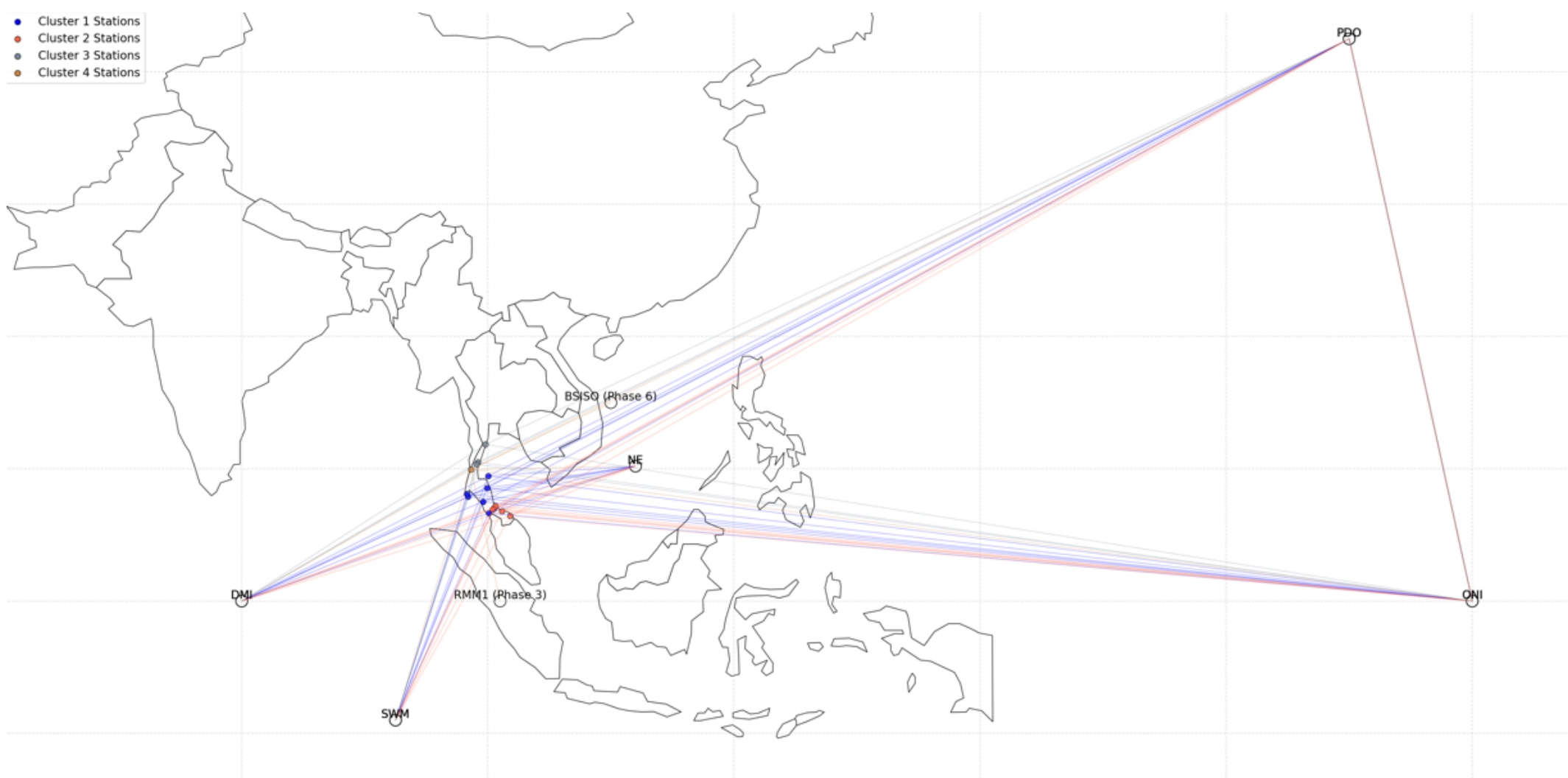

**Figure B1.** Examples of reanalysis graph-based structure of GNN models showing their stations as nodes and their teleconnections as edges in Clusters 1-4

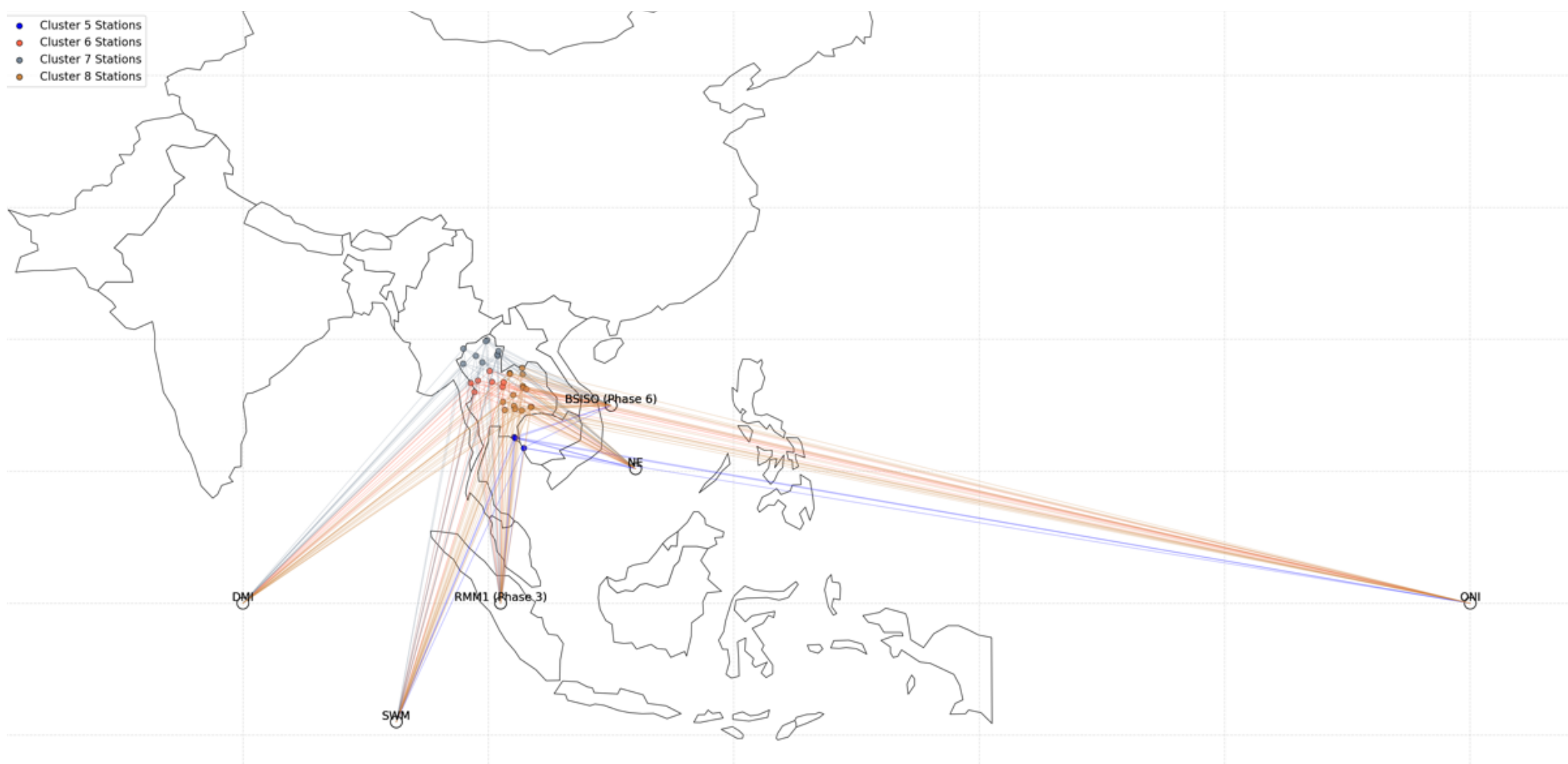

**Figure B2.** Examples of reanalysis graph-based structure of GNN models showing their stations as nodes and their teleconnections as edges in Clusters 5-8

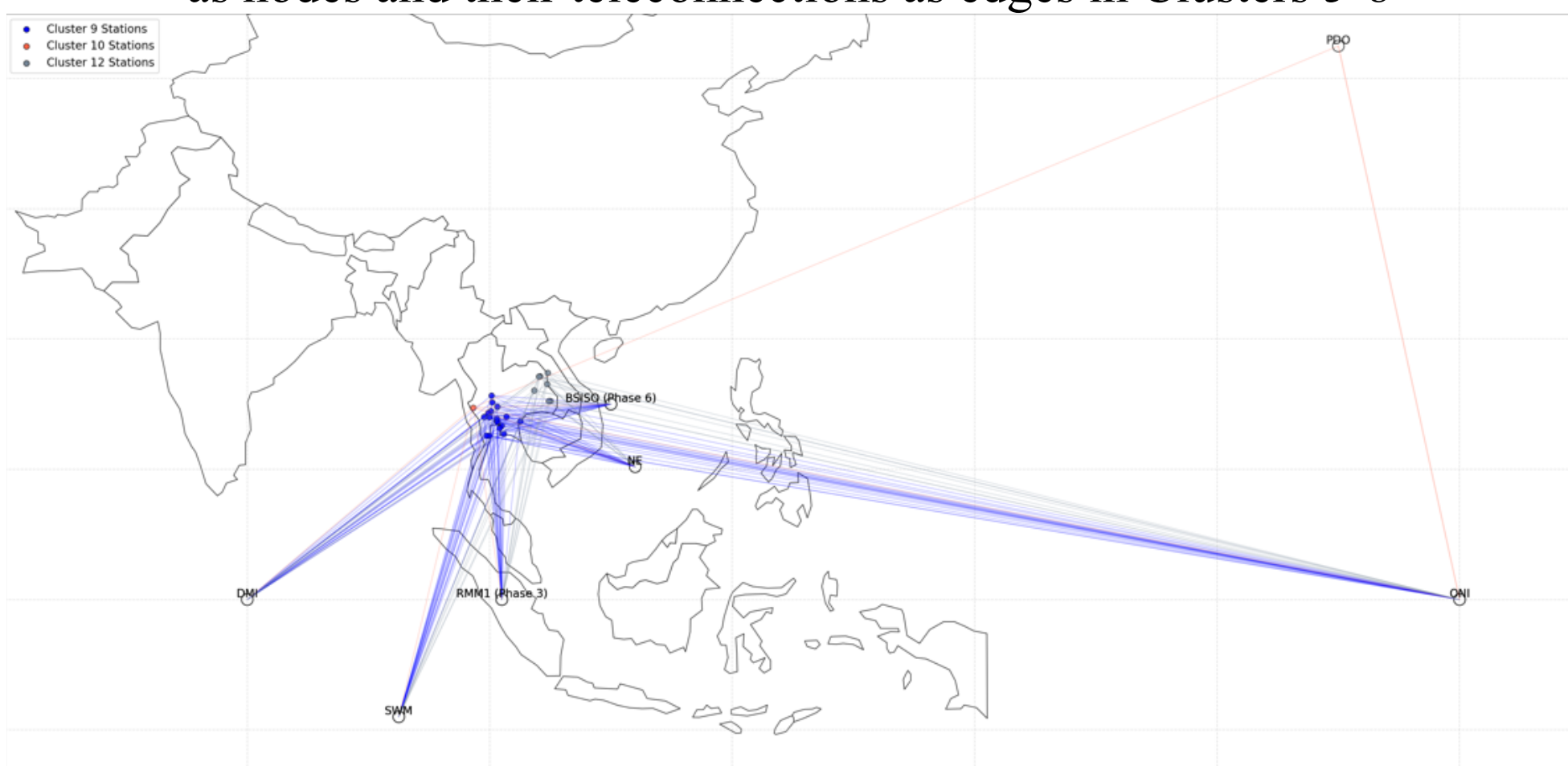

**Figure B3.** Examples of reanalysis graph-based structure of GNN models showing their stations as nodes and their teleconnections as edges in Clusters 9-12